\documentclass[journal]{./support/IEEEtran}
\IEEEoverridecommandlockouts
\usepackage{cite}
\usepackage{amsmath,amssymb,amsfonts,mathtools}
\usepackage[linesnumbered,ruled,vlined]{algorithm2e}
\usepackage{graphicx}
\usepackage{textcomp}
\usepackage{xcolor}
\usepackage{tensor}
\usepackage{float}
\usepackage{nomencl}
\makenomenclature
\usepackage{subcaption}
\usepackage[OT1]{fontenc} 
\usepackage{multirow}
\usepackage{gensymb}
\usepackage{hyperref}%

\newlength{\limage}
\newlength{\rimage}

\newlength{\aimage}
\newlength{\bimage}
\newlength{\cimage}
\newlength{\dimage}

\newlength{\textfloatsepsave} 

\def\BibTeX{{\rm B\kern-.05em{\sc i\kern-.025em b}\kern-.08em
    T\kern-.1667em\lower.7ex\hbox{E}\kern-.125emX}}
    
\DeclareMathOperator*{\argmax}{arg\,max}
\DeclareMathOperator*{\argmin}{arg\,min}

\graphicspath{{./figs/}, {./figs/gen/}}
\DeclareGraphicsExtensions{.png,.jpg,.eps,.pdf}
\title{$D^2$SLAM: Decentralized and Distributed Collaborative Visual-inertial SLAM System for Aerial Swarm}

\author{Hao Xu, Peize Liu, Xinyi Chen, Shaojie Shen
\thanks{
Manuscript was accepted by IEEE Transactions on Robotics on May 13, 2024.
This work was supported by the Research Grants Council General Research Fund (RGC GRF) project RMGS20EG20, and the HKUST-DJI Joint Innovation Laboratory.
\textit{(Corresponding author: Hao Xu.)}

All authors are with the Department of Electronic and Computer Engineering, Hong Kong University of Science and Technology, Hong Kong, China.
{\tt\small $\{$hxubc, pliuan, xchencq$\}$@connect.ust.hk, eeshaojie@ust.hk}

}
}

\begin{document}
\maketitle

\begin{abstract}
Collaborative simultaneous localization and mapping (CSLAM) is essential for autonomous aerial swarms, laying the foundation for downstream algorithms such as planning and control. To address existing CSLAM systems' limitations in relative localization accuracy, crucial for close-range UAV collaboration, this paper introduces $D^2$SLAM—a novel decentralized and distributed CSLAM system. $D^2$SLAM innovatively manages near-field estimation for precise relative state estimation in proximity and far-field estimation for consistent global trajectories. Its adaptable front-end supports both stereo and omnidirectional cameras, catering to various operational needs and overcoming field-of-view challenges in aerial swarms. Experiments demonstrate $D^2$SLAM's effectiveness in accurate ego-motion estimation, relative localization, and global consistency. Enhanced by distributed optimization algorithms, $D^2$SLAM exhibits remarkable scalability and resilience to network delays, making it well-suited for a wide range of real-world aerial swarm applications. The adaptability and proven performance of $D^2$SLAM represent a significant advancement in autonomous aerial swarm technology.
\end{abstract}


\section{Introduction}\label{sect:intro}

\IEEEPARstart{L}{ocalization} technology is critical for highly autonomous robot swarms. Unlike individual mobile robots, swarm robots are required to estimate not only their own state but also the states of other robots in the swarm. In recent years, simultaneous localization and mapping technology (SLAM) \cite{xu2022omni, tian2022kimera, lajoie2020door, choudhary2017distributed} for swarm robots, including aerial swarms \cite{zhou2022swarm,zhou2022racer}, has been greatly developed. These methods are well-known as collaborative SLAM (CSLAM) or multi-robot SLAM.

\begin{figure}[ht!]
    \centering
    \begin{subfigure}{1.0\linewidth}
        \centering
        \includegraphics[width=0.9\linewidth]{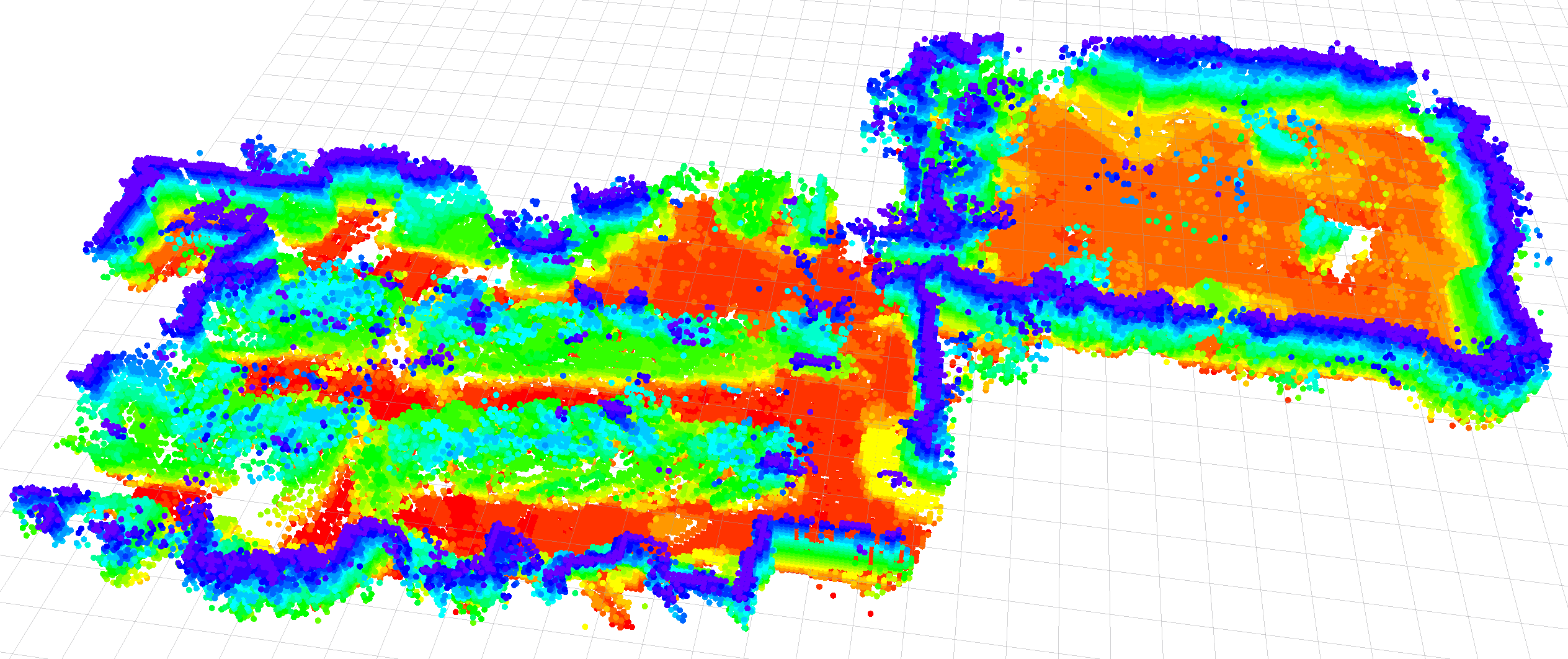}
        \caption{}\label{fig:dense_mapping_2}
    \end{subfigure}
    \begin{subfigure}{1.0\linewidth}
        \centering
        \includegraphics[width=0.9\linewidth]{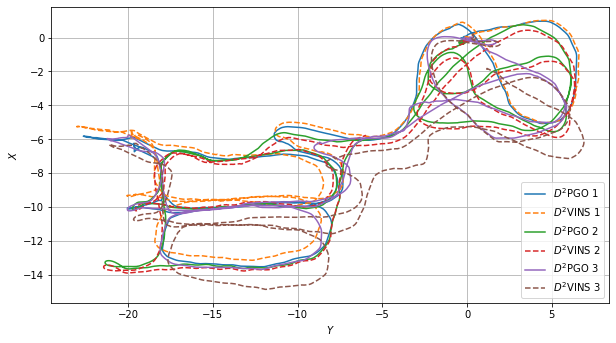}
        \caption{}\label{fig:ri_2_traj}
    \end{subfigure}
    \caption{\small{
        The demonstration of $D^2$SLAM in the HKUST RI dataset:
        a) The dense map generated by $D^2$SLAM using TSDF reconstruction, showing only the surface voxels from TSDF.
        b) The estimated trajectories of $D^2$VINS (for near-field state estimation) and $D^2$PGO (for far-field state estimation) in a three-UAV scenario.
        }}\label{fig:ri_2}
        \vspace{-0.5cm}
    \end{figure}

Considering the size, weight, and power (SWaP) constraints of aerial robot platforms, the primary goal of most SLAM algorithms on aerial robots is to provide input (such as trajectories and ESDF map) for planning and control, as shown in Fig. \ref{fig:ri_2}, rather than producing high-quality mesh.
Let's explore a few categories of practical tasks of aerial swarms to summarize the technology requirements of CSLAM.
The first category is self-assemble aerial swarm \cite{saldana2018modquad} and cooperative transportation \cite{loianno2017cooperative} using multiple UAVs, in this case, UAVs in the swarm require very precise relative localization (centimeter level) to cooperate with each other at a very close distance (usually less than a meter). The second category is inter-UAV collision avoidance \cite{zhou2020ego} and formation flight \cite{lusk2020distributed}, in this case, high accuracy relative localization (centimeter-level to decimeter-level) at a few meters' distance is required to avoid collision and maintain flight formation. Another typical task of aerial swarm is cooperative exploration of unknown space. 
In some typical task, e.g., unknown space exploration \cite{zhou2022racer}, it's essential to have a high accuracy relative localization when UAVs are near to each other to avoid inter-UAV collision for this task. However, when UAVs start to explore the unknown space and fly far away from each other, the relative localization accuracy is not important, and the global consistency of the estimated trajectories is more important to build up the global map of the unknown space.

In summary, for close-range operations, aerial swarms require high-precision relative localization. At longer distances, however, this precision becomes less critical, with a greater emphasis on the global consistency of state estimation. Additionally, accurate ego-motion estimation is vital for stable flight in all tasks.

CSLAM algorithms are influenced not only by task-specific requirements but also by communication constraints. 
Common communication modes in robot swarms typically encompass centralized systems such as WLAN, mainly used in laboratory settings, and wireless ad hoc networks. The latter, a decentralized approach, is particularly suitable for complex real-world scenarios where setting up a router for centralized communication is not feasible.
In the latter approach, typically seen in high-mobility aerial swarms \cite{xu2022omni}, robots form mesh networks with communication limited by factors like occlusion, distance, and interference. While these ad hoc networks provide effective communication at close ranges, they face challenges in bandwidth and stability over longer distances.

Moreover, the computational architecture of CSLAM may vary, primarily encompassing two major forms for robot swarms: centralized and decentralized. Centralized CSLAM processes information on a ground-station server, requiring stable network connectivity, while decentralized CSLAM \cite{lajoie2020door,xu2022omni} adapts to a broader range of environments. A well-designed decentralized approach is less dependent on stable network connectivity. 
However, decentralized methods may only partially address the issue. Some approaches, such as \cite{xu2022omni,xu2020decentralized}, are considered decentralized due to the lack of a central node, yet they replicate the same computations across multiple robots. As the number of UAVs in a swarm increases, these methods are likely to encounter computational bottlenecks.
Compared to these approaches, distributed CSLAM offers the advantage of reduced (but not cost-free) computational overhead per UAV, and enhanced privacy, as only partial information needs to be shared \cite{choudhary2017distributed}. The key advantage of distributed SLAM lies in its scalability, enabling applications in large-scale swarms.

Taking into account communication constraints and SWaP limitations, and driven by the notion that precise relative localization is essential only for nearby robots while global consistency gains importance as robots move farther apart, we propose $D^2$SLAM. This innovative system represents a novel approach to decentralized and distributed visual-inertial SLAM.
Its state estimation combines two parts: near-field and far-field state estimation for aerial swarms. Near-field state estimation in $D^2$SLAM involves estimating high-precision real-time local state (e.g., visual-inertial odometry (VIO)) and relative state between UAVs when their onboard sensors have field-of-view (FoV) overlap and good communication. Far-field state estimation involves estimating trajectories with global consistency when UAVs are far apart or in non-line-of-sight, with limited communication. These two state estimation approaches effectively address the challenges associated with state estimation in aerial swarms, as previously outlined.

The near-field estimation module in $D^2$SLAM, named $D^2$VINS (decentralized and distributed visual-inertial navigation system), is a collaborative multi-robot VIO system using sparse features. Similar to standard single-robot VIO systems like VINS-Mono \cite{qin2017vins}, $D^2$VINS maintains a local map using a sliding window and employs graph-based optimization for state estimation. 
Recent studies have shown significant advancements in applying the alternating direction method of multipliers (ADMM) \cite{shi2014linear,zhang2017distributed} to distributed bundle adjustment \cite{eriksson2016consensus, zhang2017distributed}. Inspired by these developments, we have adopted ADMM in $D^2$VINS to effectively address distributed VIO challenges.

Given visual SLAM limitations, $D^2$VINS requires overlapping sensor FoVs for accurate real-time relative state estimation at close range. Omnidirectional vision systems \cite{xu2022omni}, especially with our enhanced omnidirectional frontend, effectively address these issues in aerial swarms. With omnidirectional cameras, $D^2$VINS is not constrained by the UAVs' yaw angles, enabling precise relative localization when UAVs maintain line-of-sight, share environmental features, and communicate reliably.
Commercial stereo cameras like the Intel Realsense D435i \footnote{https://www.intelrealsense.com/depth-camera-d435i}, widely used in advanced aerial swarms \cite{zhou2022swarm,zhou2022racer}, are also supported by $D^2$SLAM. Stereo cameras, requiring less computational power, are well-suited for UAVs with limited resources. However, their optimal usage, particularly adjusting UAV yaw for FoV overlap, falls beyond the scope of this paper.

The core of far-field estimation is $D^2$PGO (decentralized and distributed pose graph optimization).
$D^2$PGO facilitates both relative and global localization through pose graph optimization, particularly when UAVs are distant or out of each other's line of sight.
Unlike $D^2$VINS, which requires good communication conditions, $D^2$PGO functions effectively even in poor communication scenarios, characterized by unstable network latency and low bandwidth.
Peng et al.'s distributed optimization method, ARock \cite{peng2016arock}, adeptly handles network latency and asynchronous updates, effectively countering the network instability and delays in pose graph optimization. Thus, we introduce an ARock-based asynchronous distributed pose graph optimization algorithm in Sect. \ref{sect:$D^2$PGO}.

The main contributions of this paper are as follows:
\begin{itemize}
\item Introduction of $D^2$SLAM, a novel decentralized and distributed SLAM system, capable of achieving high-accuracy ego-motion and relative state estimation for nearby UAVs, as well as globally consistent trajectory estimation for distant or non-visible UAVs.
\item We introduce $D^2$VINS, a distributed visual-inertial state estimator for multi-robot systems utilizing the ADMM approach. This system represents the first instance of a tightly-coupled visual-inertial odometry estimator based on distributed optimization, providing both ego-motion estimation and high-accuracy relative localization.
\item $D^2$PGO, an ARock-based, asynchronous, distributed pose graph optimization algorithm, specifically designed for multiple robots is introduced. The adaptability of $D^2$PGO to communication delays and its proficiency in managing nonlinear issues make it an ideal solution for robot swarms.  This is also the first attempt of using ARock in solving pose graph problems.
\item We conducte extensive testing of $D^2$SLAM on an aerial swarming platform, encompassing both dataset evaluations and real-world experiments. Additionally, we have open-sourced the code and custom datasets\footnote{https://github.com/HKUST-Aerial-Robotics/D2SLAM}.
\end{itemize}

\section{Related Works}
\subsection{Distributed SLAM Techniques}
This subsection explores fundamental techniques in distributed SLAM, namely distributed pose graph optimization and distributed bundle adjustment:

\subsubsection{Distributed pose graph optimization}
Pose graph optimization (PGO)\cite{Rosen2019SESync,briales2017cartan}, a technique derived from factor graph theory\cite{dellaert2017factor}, is crucial in SLAM for re-localization\cite{qin2017vins, mur2015orb} and dense mapping\cite{reijgwart2020voxgraph}. When SLAM expanded to multi-robot collaboration, PGO was naturally incorporated\cite{cunningham2010ddf, cunningham2013ddf, michael2014collaborative}. Initial PGO applications in CSLAM utilized centralized servers to solve the pose graph problem\cite{michael2014collaborative}, but faced scalability and communication challenges in swarm robot systems. Cunningham et al. proposed the first distributed SLAM method, DDF-SAM\cite{cunningham2010ddf}, using a constrained factor graph, a concept further refined in DDF-SAM2\cite{cunningham2013ddf}. Nevertheless, these methods maintain  a neighborhood graph on each agent, leading to computational inefficiencies and scalability issues.

Significant advancements in distributed pose graph optimization (DPGO) began with Choudhary et al.'s distributed Gauss-Seidel (DGS) approach\cite{choudhary2017distributed}. This method transforms the PGO problem into two linear problems, which are then distributedly solved using the Gauss-Seidel technique\cite{bertsekas1989parallel}. However, this method struggles with unfavorable convergence at high noise levels and necessitates synchronous operation. Later, the ASAPP method, based on distributed gradient descent, was proposed by Tian et al.\cite{tian2020asynchronous}, functioning as a distributed and asynchronous version of Riemann gradient descent\cite{boumal2020introduction}. This work was notable for being the first to incorporate communication considerations, introducing an asynchronous DPGO approach. Tian et al.'s subsequent work\cite{tian2021distributed} introduced the DC2-PGO method using the Distributed Riemann-Staircase, a certifiably correct DPGO approach. These DPGO methods have been incorporated into various practical SLAM systems, such as DOOR-SLAM\cite{lajoie2020door} that utilizes DGS, and Kimera-multi\cite{rosinol2020kimera, tian2022kimera} which employs both ASAPP and DC2-PGO.

\subsubsection{Distributed bundle adjustment}\label{sect:dist_ba}

Bundle adjustment (BA)\cite{triggs1999bundle}, a core component in structure-from-motion (SfM) and sparse SLAM, has evolved with significant advancements in distributed processing. Eriksson et al.\cite{eriksson2016consensus} pioneered a consensus-based distributed BA method, partitioning camera poses to enhance processing in a manner akin to the Alternating Direction Method of Multipliers (ADMM), though it requires extensive communication for landmark data. Addressing this challenge, Zhang et al.\cite{zhang2017distributed} developed a 'camera consensus' method, effectively reducing communication needs and improving convergence.
\cite{karrer2021distributed} investigates a loosely coupled approach for fusing VIO with relative measurements in distributed VIO, utilizing ARock \cite{peng2016arock}, an asynchronous distributed optimization algorithm. However, their focus on predominantly two-robot scenarios highlights the necessity for more versatile solutions. Additionally, the use of multi-state constraint Kalman filter (MSCKF) for distributed BA, as studied in \cite{zhu2021distributed, zhu2021cooperative, jung2021scalable}, although promising, may suffer from accuracy loss due to the single linearization of measurements.

Building on these insights, our approach, inspired by \cite{zhang2017distributed}, adopts a strategy of dividing landmarks into disjoint sets for an ADMM-based collaborative VIO in a distributed setting, aiming to enhance both accuracy and efficiency in multi-robot SLAM applications.
Our method stands out as the first to tightly-coupled distributed visual-inertial odometry in contrast to existing ADMM-based distributed bundle adjustment approaches. Unlike the structure-from-motion methods \cite{eriksson2016consensus, zhang2017distributed}, which don't prioritize real-time performance, our method is designed for real-time operation.
Compared to approach \cite{karrer2021distributed}, our method is applicable to multi-robot scenarios, not just limited to two UAVs. 
A unique feature of $D^2$SLAM is the directly and tightly integration of both visual and IMU measurements, a capability absent in previous works.
This enhancement maximizes the utility of visual-inertial inputs to improve accuracy, making our method highly suitable for aerial swarm applications.

\subsection{Current CSLAM Systems}
In this section, we review existing CSLAM systems, highlighting their strengths and limitations. DPGO based systems, such as \cite{lajoie2020door, tian2022kimera}, are known for their global consistency and distributed capabilities, yet they often lack in high relative localization accuracy.
Systems employing BA on landmarks, including \cite{zhu2021distributed, zhu2021cooperative, jung2021scalable, karrer2021distributed}, offer multi-camera state estimation but typically fall short in global consistency, which limits their broader application.

Additionally, methods incorporating relative measurements like UWB ranging or visual detection are gaining attention. For example, \cite{xu2020decentralized, guo2019ultra, ziegler2021distributed} apply UWB-odometry fusion for relative state estimation but face challenges in achieving global consistency. \cite{ziegler2021distributed} uses ARock for backend optimization to scale up UWB-odometry fusion, while \cite{nguyen2021range} integrates UWB anchors for enhanced global consistency, albeit with reliance on ground infrastructure. On the other hand, \cite{xu2022omni} utilizes a map-based localization approach for achieving global consistency, though UWB systems often encounter issues like obscuration and radio interference in real-world environments.

Furthermore, visual-detection-based approaches such as \cite{xu2020decentralized,xu2022omni, nguyen2019vision} face limitations due to reliance on pre-trained models, particularly when UAVs are at a distance.

Contrastingly, $D^2$SLAM not only ensures global consistency but also high relative localization accuracy at close ranges. Its distributed architecture is particularly suited for scalable swarm applications. Diverging from approaches like \cite{zhu2021distributed, zhu2021cooperative, zhou2020ego, zhou2022swarm, guo2019ultra, ziegler2021distributed, tian2022kimera} that depend on ground truth poses for initialization, $D^2$SLAM enhances system flexibility by allowing runtime merging of different robots' reference systems and sparse maps.

Unlike the distributed algorithms typically limited to dataset validation or simulation \cite{eriksson2016consensus, zhang2017distributed, zhu2021distributed, zhu2021cooperative, tian2022kimera}, our approach has undergone extensive testing both through in-the-loop flight experiments. $D^2$SLAM demonstrates promising performance in robotic systems' perception-control loops, showcasing its practical usability and effectiveness.

\section{Preliminary}
\begin{figure*}[ht]
    \centering
    \includegraphics[width=0.8\linewidth]{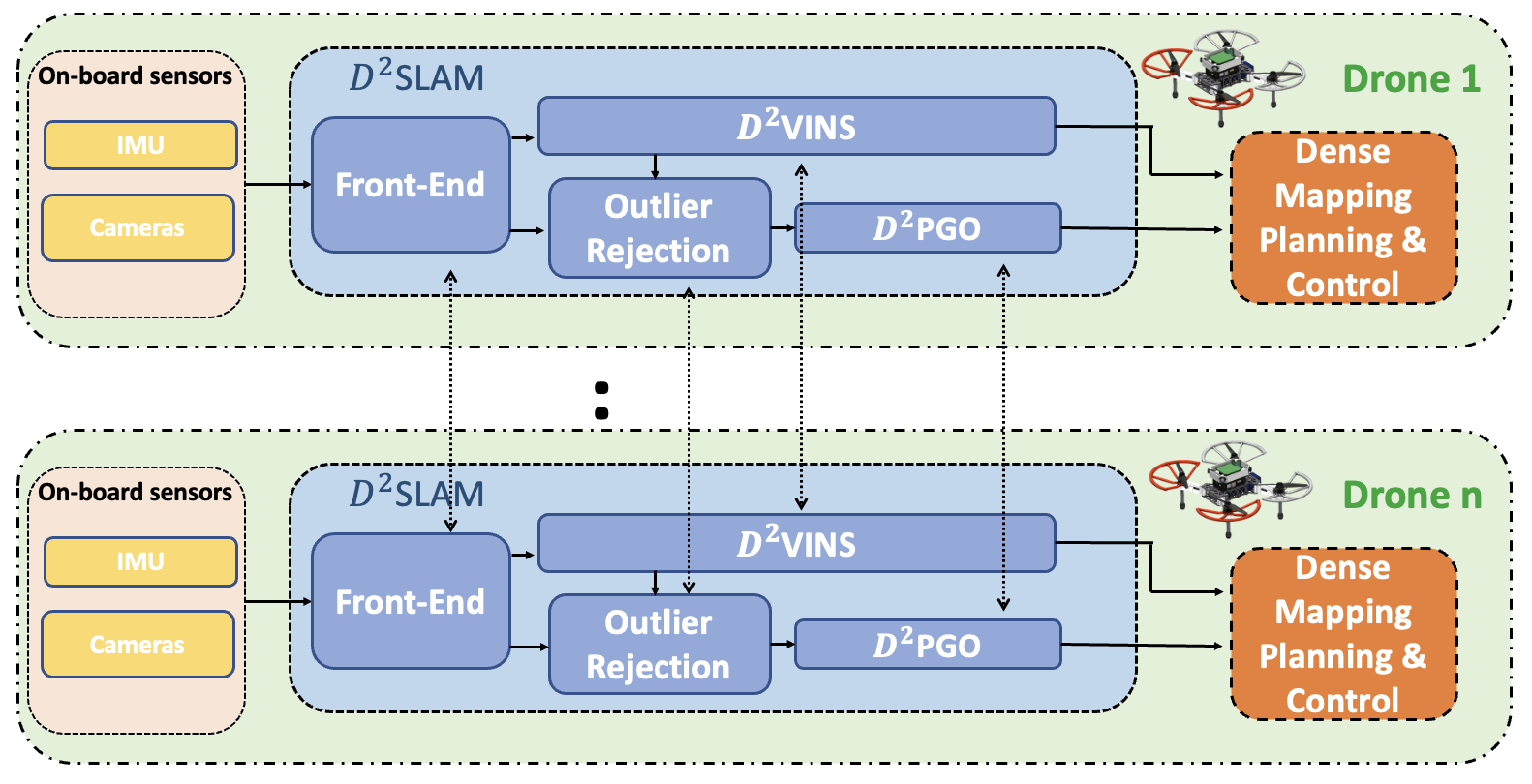}
    \caption{The architecture of $D^2$SLAM. The $D^2$SLAM is independently running on each UAV. The data will be first processed by front-end and then sent to back-end for state estimation. Results can be utilized for dense mapping, planning and control.}
    \label{fig:$D^2$SLAM-arch}
    \vspace{-0.5cm}
\end{figure*}
This paper adopts the symbol system from \cite{xu2022omni}, details of which are not repeated here.
We consider an aerial swarm consisting of $N$ UAVs, each with a unique ID, $i \in \mathcal{D}$, where $\mathcal{D} = \{1,2,\cdots,N\}$ represents the set of all UAVs in the swarm. By default, the swarm's global reference system is established based on the initial position of UAV $k$, unless stated otherwise. $D^2$SLAM operates independently on each UAV, with a focus on its implementation on the $i$-th UAV unless specified. 
Details on establishing a unified reference system among the UAVs will be provided in Sect. \ref{sect:initialization}.

\subsection{State Estimation Problem of CSLAM on Aerial Swarm}\label{subsec:def}
The state estimation problem in an aerial swarm is defined as follows: for each UAV $k\in \mathcal{D}$, estimate the 6-degree-of-freedom (DoF) pose $\tensor*[^{v_k}]{\mathbf{T}}{_i^t}$ of every other UAV $i\in \mathcal{D}^k_a$ at time $t$ in UAV $k$'s local frame, where $\mathcal{D}^k_a$ represents the set of UAVs in communication with UAV $k$. This problem is divided into two key parts for UAV $k$:
\begin{enumerate}
\item Estimation of UAV $k$'s own ego-motion state $\tensor*[]{\mathbf{\hat T}}{_k^t}$ in its local frame.
\item Estimation of the state $\tensor*[^{v_k}]{\mathbf{\hat T}}{_i^t}$ of any other UAV $i$.
\end{enumerate}

In SLAM research \cite{huang2019visual,qin2018relocalization, qiu2017model,xu2022omni}, global consistency is understood as the ability to bound the absolute trajectory error (ATE)\cite{Zhang18iros}, ensuring that state estimation errors do not progressively increase with the robot's movement.
In this paper, it means that the estimated poses $\tensor*[^{v_k}]{\mathbf{\hat T}}{_i^t}$ remain accurate and do not suffer from drift as the robot moves.

\subsection{Decentralized vs Distributed}
\begin{figure}[ht]
    \centering
    \includegraphics[width=0.8\linewidth]{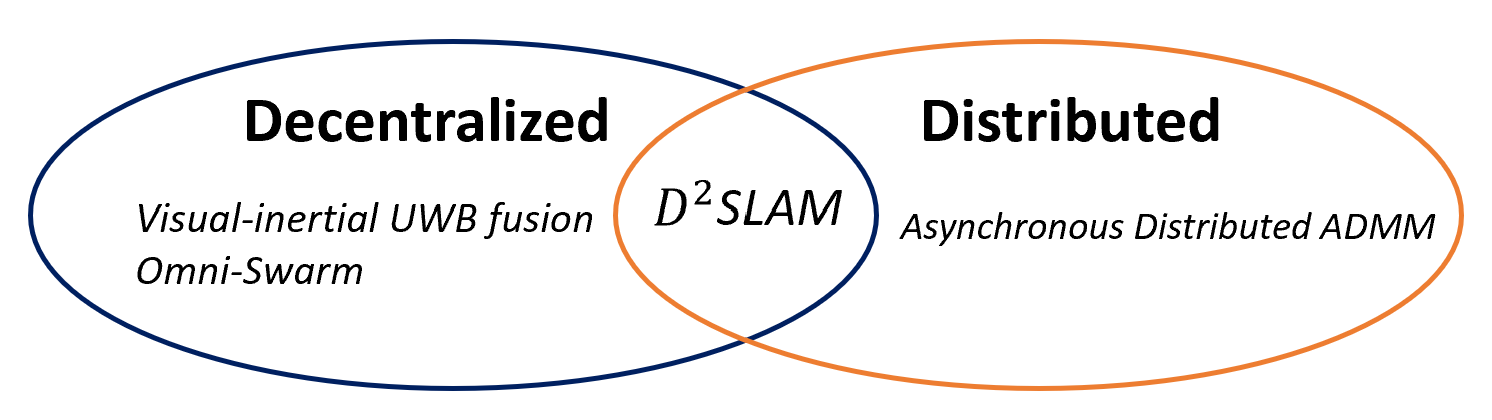}
\caption{
    \small{Visual-inertial UWB fusion \cite{xu2020decentralized} and Omni-Swarm \cite{xu2022omni} are decentralized,
    and asynchronous Distributed ADMM \cite{zhang2014asynchronous} is distributed. $D^2$SLAM is both decentralized and distributed.}
    }\label{fig:d2}
    \vspace{-0.5cm}
\end{figure}

Decentralized and distributed systems, often used interchangeably, have distinct meanings. Decentralized systems operate without a central node, with algorithms running independently across multiple nodes. In contrast, distributed systems, as defined in \cite{van2002distributed}, spread computation or resources like storage across multiple nodes, aiming for non-redundant processing. However, decentralized systems can involve significant redundant computation, while distributed systems may still rely on a central node for coordination.
The relationship between decentralization and distribution is illustrated in Fig. \ref{fig:d2}. For instance, Omni-swarm \cite{xu2020decentralized} is decentralized but not distributed, lacking a central node but involving redundant optimization across the swarm, leading to computational inefficiency. Conversely, Asynchronous Distributed ADMM \cite{zhang2014asynchronous} employs a master-worker structure that distributes computation across multiple nodes, yet relies on a central node.

This paper aims to develop a fully distributed and decentralized ($D^2$) CSLAM algorithm, $D^2$SLAM.
This approach ensures robustness against single node failures and communication losses, enabling the swarm to dynamically form sub-swarms or integrate into a larger swarm.
Moreover, $D^2$SLAM efficiently utilizes the computational resources of each UAV, enhancing overall system performance.

\section{System Overview}

\subsection{$D^2$SLAM System Architecture}
Fig. \ref{fig:$D^2$SLAM-arch} depicts the architecture of $D^2$SLAM. On each UAV, the front-end processes visual inputs for key frame extraction, sparse feature tracking, and loop closure detection. These processed data are then forwarded to the back-end for two types of state estimations: collaborative visual odometry (near-field state estimation) and pose graph optimization (far-field state estimation), handled by $D^2$VINS and $D^2$PGO, respectively. The resulting estimations support further processes like dense map building, planning, and control.

\subsection{Communication Modes}\label{sect:comm_strategy}
$D^2$SLAM employs a combination of three distinct communication modes to enhance system robustness and minimize communication overhead:
\begin{itemize}
\item \textbf{Discover mode}: Activated when UAVs' relative states are unknown, this mode involves broadcasting complete keyframes (Sect. \ref{sect:preprocessing}) and states for both near (Sect. \ref{sect:d2vins}) and far-field (Sect. \ref{sect:$D^2$PGO}) state estimation. This facilitates rapid initialization of relative state estimation and supports map merging (Sect. \ref{sect:initialization}) for integrating UAVs' reference systems.
\item \textbf{Near mode}: When other UAVs are predicted to be in proximity (within 3 to 5 meters), based on current state estimates, $D^2$SLAM broadcasts complete keyframes and states necessary for near and far-field state estimation. The proximity prediction relies on state estimation from both $D^2$PGO and $D^2$VINS.
\item \textbf{Far mode}: Used when UAVs are distant from each other, this mode transmits compact keyframes (Sect. \ref{sect:preprocessing}) and performs distributed loop closure detection (Sect. \ref{sect:dist_loop}). It broadcasts states for far-field state estimation. In far mode, near-field state estimation simplifies to single-robot VIO.
\end{itemize}

Fig. \ref{fig:comm_state_machine} depicts $D^2$SLAM's communication mode transitions.
After initialization, near-field state estimation is limited to near mode to save bandwidth, while far-field state estimation operates in all modes for global consistency. Details on data transmission per mode are provided in later sections.

\subsection{Multi-Robot Map Merging}\label{sect:initialization}
\begin{figure}[h!]
    \centering
    \begin{subfigure}{1.0\linewidth}
        \centering
        \includegraphics[width=1.0\linewidth]{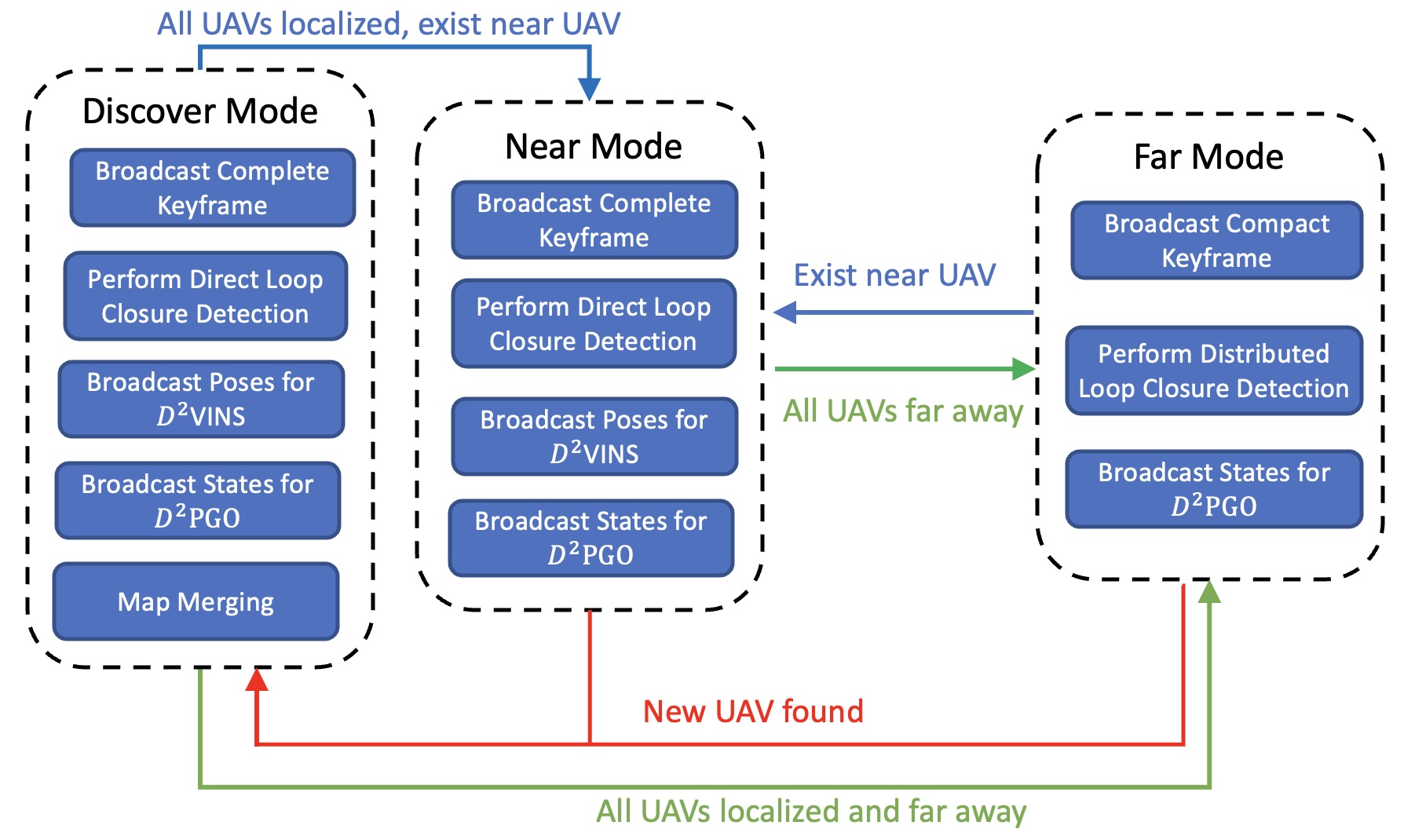}
        \caption{}\label{fig:comm_state_machine}
    \end{subfigure}
    \begin{subfigure}{1.0\linewidth}
        \centering
        \includegraphics[width=1.0\linewidth]{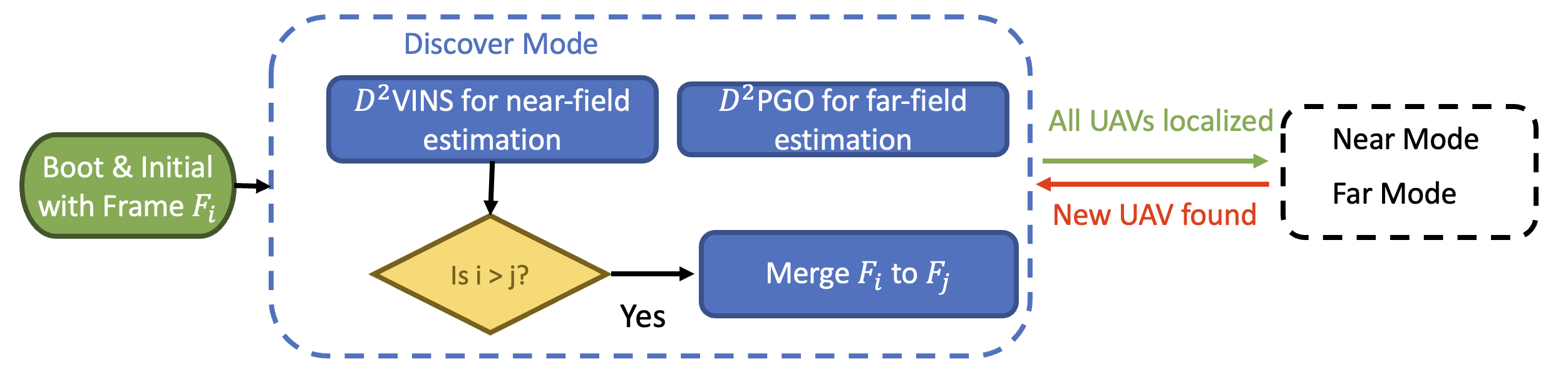}
        \caption{}\label{fig:mapmerge_new}
    \end{subfigure}
    \caption{\small{
        a) A state machine governing $D^2$SLAM's communication modes begins in discover mode. In this initial phase, each UAV broadcasts its complete keyframe to support rapid initialization and transmits data essential for $D^2$VINS. Once UAVs have successfully initialized their relative states, $D^2$SLAM transitions to either near or far mode, depending on their estimated relative positions. The system can fluidly alternate between near and far modes in response to changes in UAV positioning. Importantly, if a new UAV enters the system, typically seen during the initialization phase, $D^2$SLAM reverts to discover mode to incorporate this new member.
        b) Map merging occurs during the discover mode, as shown in the flow charts. If near or far-field state estimation identifies other UAVs, the UAV with the higher index transfers its map state to the UAV with the lower index.
       }}\label{fig:map_merging}
       \vspace{-0.5cm}
\end{figure}

In $D^2$SLAM, each robot starts with its own local reference frame $\mathbf{F}_i$ at boot time. When map-merging occurs, typically during encounters with other UAVs or revisiting areas covered by others, the UAVs align their coordinate systems to form a unified reference system. This alignment, triggered by relative measurements, integrates the coordinate systems without exchanging landmarks, allowing each UAV to maintain its individual sparse map. Specifically, the system adopts the reference frame of the UAV with the smaller ID, and the larger ID UAV's states in $D^2$VINS and $D^2$PGO are converted to this unified system.

Fig. \ref{fig:mapmerge_new} illustrates the map-merging process in $D^2$SLAM. UAVs continuously transmit heartbeat packets for detection by others, facilitating their discovery. During the discover mode, $D^2$SLAM broadcasts information until the swarm's UAVs are unified under a single coordinate system. After completing map-merging, subsequent state estimations proceed without this process, unless new UAVs join the swarm.

\subsection{Operating Conditions of $D^2$SLAM}\label{sect:operating_range}

$D^2$SLAM adapts to environmental and communication constraints by transitioning to single-robot SLAM or VIO when necessary. This ensures aerial swarm stability and enables short-term formation flight using only VIO, as shown in \cite{lusk2020distributed}. Detailed discussions on these limitations follow.

\subsubsection{Communication} 

In $D^2$SLAM, a loss of communication triggers a transition to single-robot SLAM, referred to as the near communication mode. Near-field state estimation requires extensive data exchange and real-time performance, hence a robust, low-latency network is essential. Conversely, far-field state estimation can function effectively with lower bandwidth and some latency.

\subsubsection{Environments}

In open environments with limited environmental features, such as grasslands or rough walls, sparse visual SLAM faces challenges in feature matching for relative localization and loop closure detection. 
In response to these limitations, our system is designed to downgrade to single-robot VIO, ensuring flight safety under such conditions.

\subsubsection{Field of View} 
Accurate relative localization in $D^2$SLAM is contingent on FoV overlap and a sufficient number of common features.
Yaw alignment is crucial for FoV-limited cameras, but this requirement is eliminated with omnidirectional cameras.
The local precision of $D^2$VINS helps offset brief losses of common features, ensuring reliable relative state estimation.
When UAVs lose shared vision with others for an extended period, $D^2$VINS degrades to single-robot VIO.
We believe that the need for accurate relative localization lessens when UAVs are either out of sight for extended periods or too distant to share features.

\section{Front-end}
\begin{figure*}[h]
    \centering
    \includegraphics[width=0.8\linewidth]{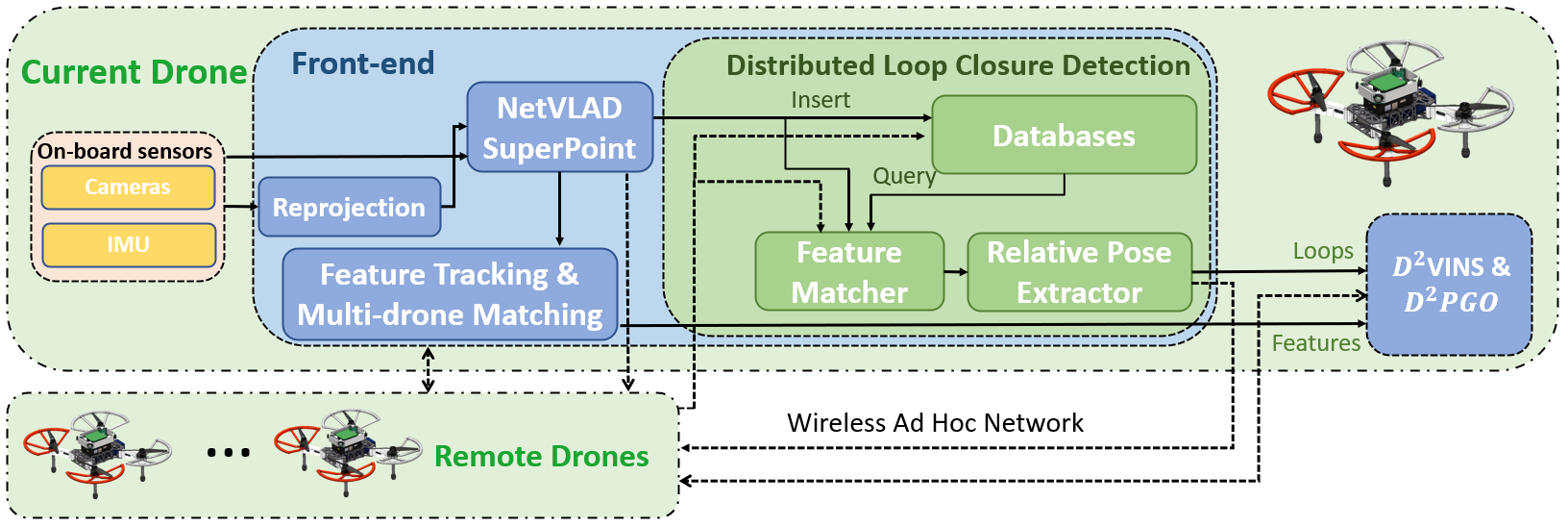}
    \caption{\small{The front-end of $D^2$SLAM processes visual data in several stages. Initially, the data undergoes reprojection, followed by extraction of global descriptors and features. Subsequently, it is used for feature tracking, multi-UAV feature matching, and loop closure detection. The final results are then fused in the back-end.}}
    \label{fig:frontend}
\end{figure*}

$D^2$SLAM features a versatile and efficient front-end that supports various camera configurations, including omnidirectional and stereo cameras.

\subsubsection*{Structure of front-end}
Fig. \ref{fig:frontend} illustrates the components of our front-end, which include:
1) Pre-processing fisheye images from omnidirectional cameras;
2) General pre-processing for extracting landmarks and global descriptors;
3) Feature tracking for the local UAV;
4) Sparse feature matching across multiple UAVs;
5) Distributed loop closure detection.

\subsubsection*{Landmarks \& features}
Similar to other sparse visual SLAM systems, $D^2$SLAM utilizes landmark points to model the external environment.
It treats a set of consistently tracked or matched 2D feature points across images as measurements of identical landmarks. 
Subsequent sections will detail the methods for tracking and matching landmark-related features, both by individual UAVs and collaboratively among multiple UAVs.

\subsection{Visual Data Preprocessing} \label{sect:preprocessing}
\begin{figure*}[h]
    \centering
    \begin{subfigure}{\textwidth}
        \centering
        \includegraphics[width=0.8\linewidth]{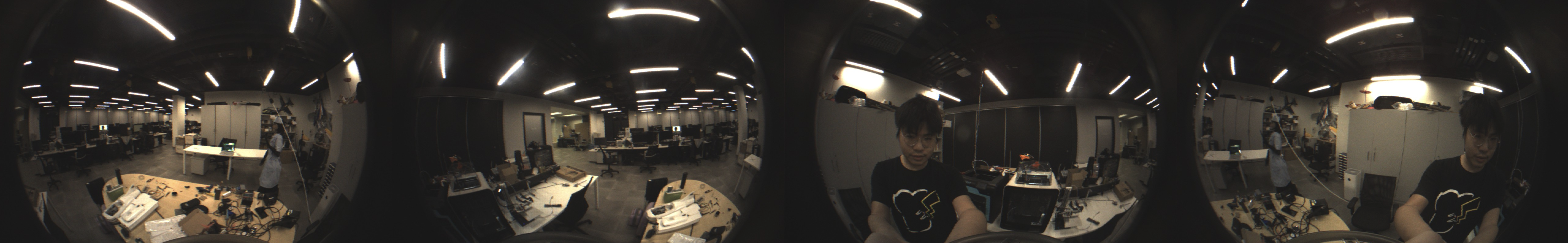}
        \caption{}\label{fig:fisheye_raw_d2}
    \end{subfigure}
    \begin{subfigure}{\textwidth}
        \centering
        \includegraphics[width=0.8\linewidth]{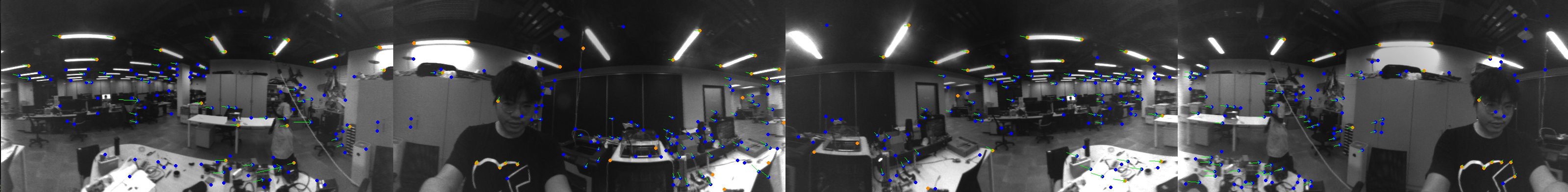}
        \caption{}\label{fig:fisheye_feature_tracking}
    \end{subfigure}
    
    \caption{\small{Fisheye Image Demonstration: (a) Original fisheye images; (b) Reprojected fisheye images featuring tracking. Blue points represent SuperPoint features, while orange points indicate LK features, and green arrows depict the inter-frame movements of these features.}}\label{fig:fisheye}
\end{figure*}
\begin{figure}[h]
    \centering
    \includegraphics[width=0.6\linewidth]{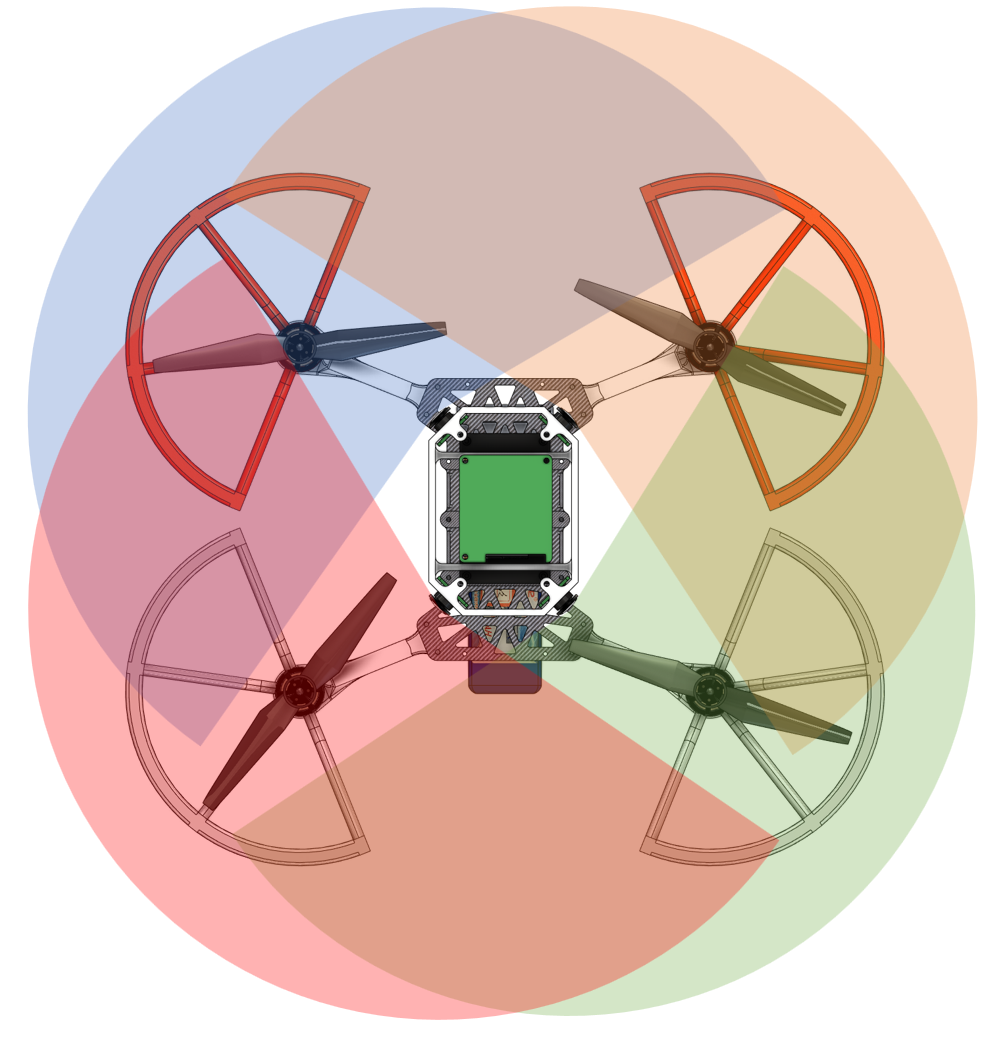}
    \caption{\small{Omnidirectional Camera Configuration: The transparent areas represent the Field of View (FoV) of each camera. These cameras boast a lateral FoV of approximately 200 degrees and feature an overlap of 110 degrees between adjacent cameras.
    }}\label{fig:fisheye_config}
    \vspace{-0.5cm}
\end{figure}

In this work, as depicted in Fig. \ref{fig:fisheye_config}, we employ a quad fisheye camera system as an omnidirectional camera. This configuration, compared to dual fisheye systems \cite{xu2022omni, gao2017dual}, offers more effective FoV usage and generates superior quality depth information. Its advantages have led to its adoption in both SLAM research \cite{won2020omnislam} and advanced commercial UAV products, such as the DJI Mavic 3 \footnote{https://www.dji.com/mavic-3}.

Fisheye cameras produce significantly distorted images, challenging direct application of conventional vision techniques, as shown in Fig. \ref{fig:fisheye_raw_d2}.
Following \cite{plaut20213d}, we first reproject these images into a cylindrical projection (see Fig. \ref{fig:fisheye_feature_tracking}) before processing. \cite{plaut20213d} demonstrates that applying CNNs trained on public datasets to these reprojected images yields effective results.

In $D^2$SLAM, to conserve communication bandwidth, complete keyframe information, such as SuperPoint features and global descriptors for each camera view, is broadcasted to other UAVs only in discover or near communication modes. On the other hand, the compact keyframes, include only the NetVLAD descriptors, are used in far mode to save bandwidth.

\subsection{Descriptor Extraction \& Dimension Reduction}
We employ MobileNetVLAD \cite{sarlin2019coarse} to construct a global descriptor for each camera view, and SuperPoint \cite{detone2018superpoint} for landmark extraction, coupled with PCA for dimension reduction of the descriptors. These techniques, validated in our previous work \cite{xu2022omni}, are not elaborated here for brevity.

\subsection{Sparse Feature Tracking} \label{sect:feature_tracking}
Sparse visual SLAM typically employs two types of feature tracking: 1) Lucas-Kanade (LK) method, as in VINS-Mono \cite{qin2017vins}, offering robustness in feature-poor environments like grass, and 2) Descriptor-based matching, notably in ORB-SLAM \cite{murORB2}, with CNN-based approaches excelling in handling large parallax for multi-UAV matching and loop closure detection, though less effective in feature-sparse areas. Given the importance of both multi-UAV feature matching and robustness in our work, we adopt a hybrid front-end to meet these requirements, illustrated in Fig. \ref{fig:fisheye_feature_tracking}.

Each camera view in $D^2$SLAM tracks up to $N_{max}$ (typically 100-200) landmarks. Initially, we extract $N_{max_{sp}}$ (50-150) SuperPoint \cite{detone2018superpoint} sparse features and their descriptors, assuming $N_{sp}$ successful extractions. SuperPoint is preferred for its robustness over ORB features \cite{murORB2}. These landmarks are tracked inter-frame using a kNN matcher with ratio test \cite{lowe2004distinctive}. Additionally, Shi-Tomasi corners are detected to fill the remaining quota up to $N_{max}$ and tracked with the LK method. SuperPoint features are utilized for loop closure detection and multi-UAV matching, while LK landmarks aid ego-motion estimation. In feature-scarce environments, if no SuperPoint features are detected ($N_{sp}=0$), the system defaults to single-robot VIO. Similiar to \cite{qin2017vins}, keyframe determination is based on parallax and the count of new landmarks after feature tracking.

Upon completing per-frame tracking, multi-view feature tracking is applied across different cameras on the same UAV. For stereo inputs, matching SuperPoint features and LK optical flow between left and right cameras is straightforward. In contrast, with quad fisheye cameras, as illustrated in Fig. \ref{fig:fisheye_config}, each camera shares landmarks only within the half-plane of its neighboring camera. Therefore, feature matching for SuperPoint features is limited to these corresponding half-planes.
Landmarks' positions are predicted based on the cameras' internal and external parameters, serving as initial values for solving the LK optical flow.

\subsection{Multi-UAV Sparse Feature Matching}\label{sect:multi-UAV-match}
When a new keyframe is received, the system first searches the sliding window for the most recently matching keyframe, determined by their global descriptors' inner product surpassing the threshold $\tau_{mg}$ (default 0.8). Following a successful match, kNN feature matching with ratio test is performed on the Superpoint features of these keyframes to establish feature correspondences. Subsequently, once a match is confirmed, we combine the corresponding landmark points of each matched feature pair into an inter-UAV landmark.

\subsection{Distributed Loop Closure Detection}\label{sect:dist_loop}
\begin{figure}
    \centering
    \includegraphics[width=1.0\linewidth]{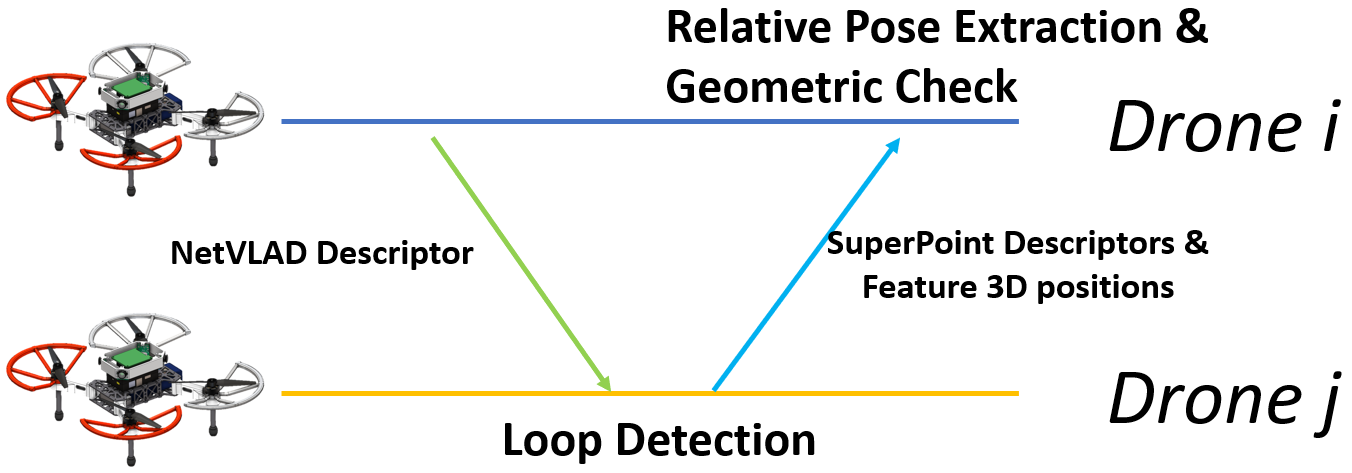}
    \caption{\small{Demonstration of distributed loop closure detection: UAVs transmit compact keyframes, which include NetVLAD global descriptors, across the swarm. Upon receiving this data from UAV \textit{i}, UAV \textit{j} initiates loop closure detection. In case of a successful match, UAV \textit{j} sends a complete keyframe containing landmark information back to UAV \textit{i}, enabling the extraction of the final relative pose.
    }}\label{fig:dist_loop}
    \vspace{-0.5cm}
\end{figure}

Following \cite{xu2022omni}, we utilize Faiss \cite{johnson2019billion} for whole image searching, kNN for feature matching, Perspective-n-Point (PnP) for relative pose extraction, and implement outlier rejection. Subsequently, loop edges are processed by the backend for pose graph optimization. 
As illustrated in Fig. \ref{fig:dist_loop}, $D^2$SLAM adopts distributed loop closure detection \cite{lajoie2020door, tian2022kimera} to minimize communication bandwidth in far mode, and direct loop clousure detection \cite{xu2022omni} in near and discover mode.

Unlike \cite{xu2022omni}, in $D^2$SLAM, we use 3D positions of Superpoint landmarks, estimated by $D^2$VINS from historical frames, along with 2D observations from the current frame for PnP computation.
Additionally, with quad fisheye cameras as input, solving a multi-camera PnP problem becomes necessary for relative pose extraction and geometric verification.
In practice, we initially apply the UPnP RANSAC algorithm \cite{kneip2014upnp} for pose estimation and outlier rejection. Subsequent inlier results are then used in a bundle adjustment problem for multi-camera PnP \cite{kneip2014opengv}, enhancing pose accuracy.

\section{Near-field state estimation: $D^2$VINS} \label{sect:d2vins}
\begin{figure}[t]
    \centering
    \includegraphics[width=1.0\linewidth]{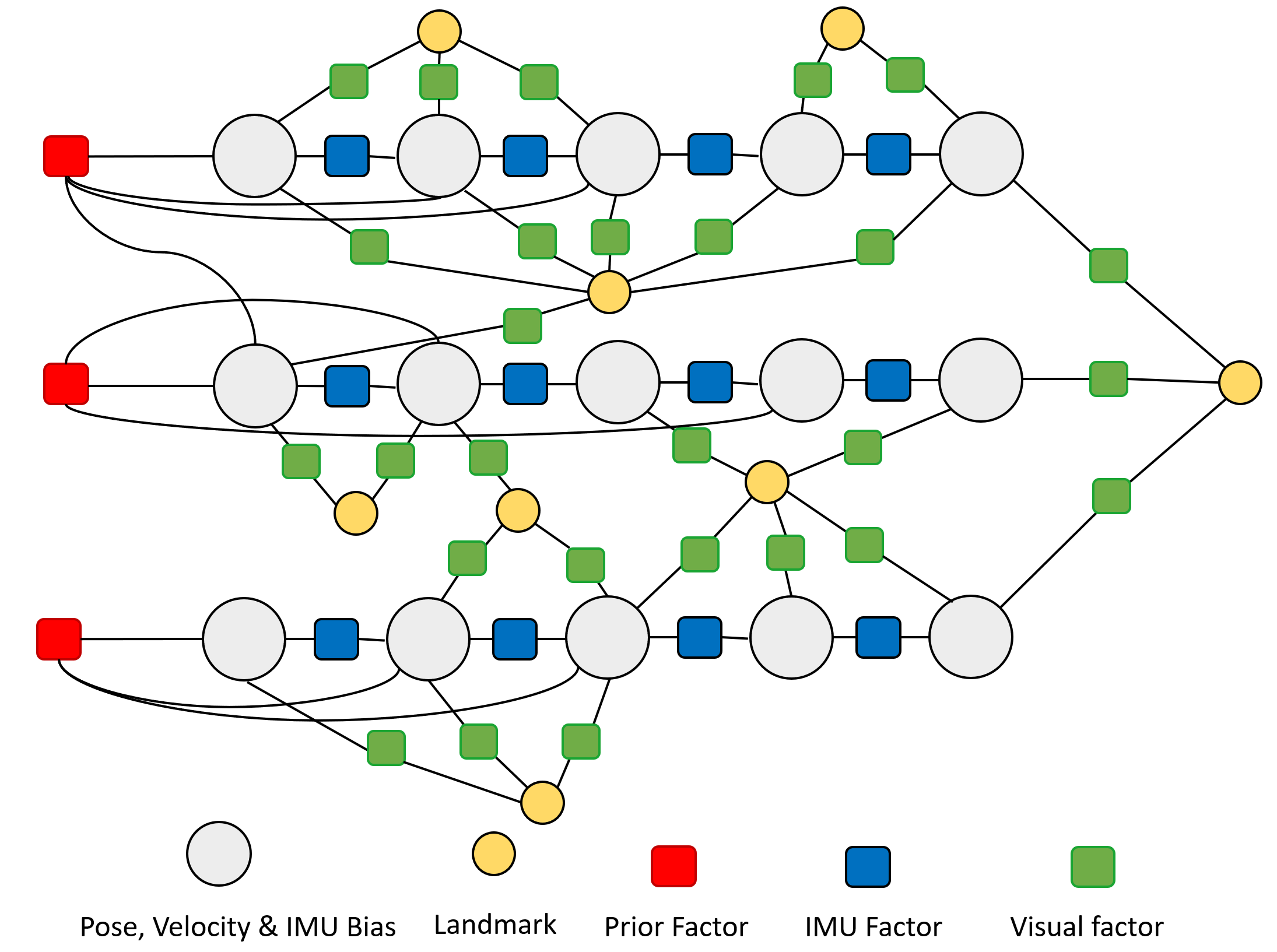}
    \caption{\small{A demonstration of the factor graph in $D^2$VINS. The states within the sliding window, comprising pose, velocity, and IMU bias, are interconnected by various factors including IMU, visual, and prior factors.    }} \label{fig:factor_graph}
    \vspace{-0.5cm}
\end{figure}

\subsection{Problem Formulation}
To facilitate near-field state estimation in $D^2$SLAM, we introduce $D^2$VINS, a collaborative visual-inertial odometry. Designed for real-time, accurate 6-DoF odometry estimation of both the local UAV and nearby UAVs, $D^2$VINS operates on a graph optimization framework with a sliding window, akin to \cite{qin2017vins}. Its state estimation problem is modeled using a factor graph \cite{dellaert2017factor}, depicted in Fig. \ref{fig:factor_graph}. State estimation within the sliding window is achieved by maximizing the a posteriori (MAP) inference of this factor graph through distributed non-linear least squares optimization.
The state of the collaborative visual-inertial-odometry problem is defined as:
\begin{equation}
    \begin{aligned}
        \mathcal{X} = \Big[& ^{v_k}\mathbf{\tilde T}_{0}^{t_0},  ^{v_k}\mathbf{\tilde T}_{0}^{t_1}, \cdots ^{v_k}\mathbf{\tilde T}_{0}^{t_{m - 1}},
        \mathbf{b}_0^{t_0}, \mathbf{b}_0^{t_1}, \cdots \mathbf{b}_0^{t_{m - 1}},\\
        &\cdots\ ^{v_k}\mathbf{\tilde T}_{n-1}^{t_0}, \ ^{v_k}\mathbf{\tilde T}_{n-1}^{t_1},\cdots ^{v_k}\mathbf{\tilde T}_{n-1}^{t_{m - 1}}, \mathbf{b}_{n-1}^{t_0}, \mathbf{b}_{n-1}^{t_1}, \cdots\ \\ 
        &\mathbf{b}_{n-1}^{t_{m - 1}}, l_0, l_1, l_2,\cdots l_L\Big]^T,
    \end{aligned}
\end{equation}
where $\tensor*[^{v_k}]{\mathbf{\tilde T}}{_i^{t_j}}$ is the estimated pose of UAV $i$ at time $t_j$ in the local frame of UAV $k$, $m$ is the length of the sliding window and $n$ is the number of the UAVs in the swarm,
$\mathbf{b}_i^{t_j} = [\tensor*[^{v_k}]{\mathbf{v}}{_i^{t_j}}, b_a^{t_j}, b_g^{t_j} ]$ is the velocity, acceleration bias and angular velocity bias of the IMU of UAV $i$ at time $t_j$,
$L$ is the total number of the landmark.
In this paper, we use the inverse depth parametrization for landmarks \cite{civera2008inverse}, where 
$l_j$ represents the inverse of the distance from landmark $j$ to the keyframe it is associated with.
For convenience, we denote the set of all poses in $\mathcal{X}$ as $\mathcal{T}$.
The \textit{optimization problem for collaborative visual-inertial-odometry} is formulated as,
\begin{equation} \label{eq:$D^2$VINS_full}
    \begin{aligned}
        \min_{\mathcal{X}} \Bigg\{ \left\Vert \mathbf{r}_p - \mathbf{H}_p \mathcal{X} \right\Vert^2 + 
        &\sum_{k\in\mathcal{B}}{\left\Vert r_\mathcal{B}(\tilde z_{b_k}^{b_{k+1}}, \mathcal{X}) \right\Vert_{\Sigma_\mathcal{B}}} +  \\
        &\sum_{k\in\mathcal{L}}{\rho\left(\left\Vert r_\mathcal{L}(\tilde z_{l_k}, \mathcal{X}) \right\Vert_{\Sigma_\mathcal{L}}\right) }
        \Bigg\}, \\
        &s.t.\ \mathbf{P}_i \in \mathbf{SE}(3),\ \forall \mathbf{P}_i \in \mathcal{T}
    \end{aligned}
\end{equation}
where $\rho(\cdot)$ is the Huber norm \cite{huber1992robust},
$\left\Vert(\cdot)\right\Vert_\Sigma$ is the Mahalanobis norm,
$\mathcal{B}$ is the set of IMU factors,
$\mathbf{r}_\mathcal{B}$ is the residual of the IMU measurement factor,
$\mathcal{L}$ is the set of visual measurements,
and $\mathbf{r}_\mathcal{L}$ is the visual measurement residual, $\mathbf{r}_\mathcal{B}$ and $\mathbf{r}_\mathcal{L}$ are define in \cite{qin2017vins}; $\left\Vert \mathbf{r}_p - \mathbf{H}_p \mathcal{X} \right\Vert^2$ is the prior factor generated from previous marginalization, which will be detailed in Sect. \ref{sect:margin}.

\subsection{Decentralized and Distributed Optimization of Collaborative visual-inertial-odometry problem}
\subsubsection{Alternating Direction Method of Multipliers (ADMM) algorithm}
In this section, the target of decentralized and distributed optimization is to solve the problem,
\begin{equation}\label{eq:dist_problem}
    \begin{aligned}
        \sum_{i=1}^{n} &f_i(\mathbf{x}_i), \\
        & s.t.\ \mathbf{x}_i = \mathbf{z},
    \end{aligned}
\end{equation}
using decentralized (and distributed) ADMM method \cite{shi2014linear,zhang2017distributed}, 
where $\mathbf{x}_i$ is the local state for UAV $i$ and $z$ is the global state, Problem (\ref{eq:dist_problem}) can be jointly solved by iteratively performing the following update on each UAV $i$,
\begin{align}
\mathbf{x}_i^{t+1} &= argmin_{\mathbf{x}_i}{\left\{f_i(\mathbf{x}_i) + \frac{\rho}{2} \Vert \mathbf{x}_i - (\mathbf{z}^{t} - \mathbf{u}_i^{t}) \Vert^2\right\}} \label{eq:dist_update_1}\\
\mathbf{z}^{t+1} &= \frac{1}{n} \sum_{i=1}^{N} \mathbf{x}_i^{t+1} \\
\mathbf{u}_i^{t+1} &= \mathbf{u}_i^t + \mathbf{x}_i^{t+1} - \mathbf{z}^t\label{eq:dist_update},
\end{align}
Here, in each iteration, each UAV solves Problem (\ref{eq:dist_update_1}) and updates its states by averaging the results obtained from all UAVs.

\subsubsection{Problem splitting} \label{sect:partition}
As outlined in Sect. \ref{sect:dist_ba}, effectively partitioning the problem and state is crucial in ADMM optimization. In $D^2$VINS, we split the landmarks $\mathcal{L}$ into disjoint sets ${\mathcal{L}_1, \mathcal{L}_2, \cdots \mathcal{L}_N}$, where $\mathcal{L}_i$ represents the set allocated to UAV $i$. This partitioning approach and the landmark distribution policy are further elaborated in Sect. \ref{sect:lmanager}.Additionally, we denote the set of poses observing landmarks in $\mathcal{L}_i$ as $\mathcal{T}_i$.
Consequently, we define the local state $\mathcal{X}_i$ of UAV $i$ as,
\begin{equation}
    \mathcal{X}_i = \left[\mathcal{T}_i, \mathcal{L}_i, \mathbf{b}_i^{t_0}, \mathbf{b}_i^{t_1} \cdots \mathbf{b}_i^{t_{m-1}} \right]^T,
\end{equation}
where $ \mathbf{b}_i^{t_0}, \mathbf{b}_i^{t_1} \cdots \mathbf{b}_i^{t_{m-1}}$ are the state of the velocity and IMU bias of UAV $i$, they are only essential for local optimization.
Due to the presence of the hybrid tracking feature method (Sect. \ref{sect:feature_tracking}), $\mathcal{X}_i$ necessarily contains all i's pose in $\mathcal{X}$.
This ensures that once all remote UAV are disconnected, $D^2$VINS can still estimate the ego-motion, i.e., degrade to a single-robot VIO.

\subsubsection{Distributed optimization for $D^2$VINS}\label{sect:dist_opti_vins}
Rewriten Eq. (\ref{eq:$D^2$VINS_full}) to the form of decentralized optimization (\ref{eq:dist_problem}), we have,
\begin{equation} \label{eq:$D^2$VINS_dist}
    \begin{aligned}
        \min_{ \mathcal{X}_i} & \sum_{i=1}^{N} f_{cvio_i} (\mathcal{X}_i) & \\
        &  f_{cvio_i} (\mathcal{X}_i) =  \left\Vert \mathbf{r}_{p_i} - \mathbf{H}_{p_i} \mathcal{X}_i \right\Vert^2 + 
        \sum_{k\in\mathcal{B}}{\left\Vert r_{\mathcal{B}_i}(\tilde z_{b_k}^{b_{k+1}}, \mathcal{X}_i) \right\Vert_{\Sigma_{\mathcal{B}_i}}} \\ 
        & + \sum_{k\in\mathcal{L}}{\rho\left(\left\Vert r_\mathcal{L}(\tilde z_{l_k}, \mathcal{X}_i) \right\Vert_{\Sigma_\mathcal{L}}\right) } \\
        &s.t.\ \mathcal{X}_i = \mathbf{E}_i \mathcal{X}, \\
        &s.t.\ \mathbf{P}_j \in \mathbf{SE}(3),\ \forall \mathbf{P}_j \in \mathcal{T}_i 
    \end{aligned}
\end{equation}
where $\mathcal{B}_i$ is the set of the IMU factors of UAV $i$, $\mathbf{P}_j$ is the $j$-th pose in $\mathcal{T}_i$, $\mathbf{E}_i$ project the global state $\mathcal{X}$ to local state $\mathcal{X}_i$,
$f_{cvio_i}(\mathcal{X}_i)$ is the subproblem of the global optimization Problem (\ref{eq:$D^2$VINS_full}) on UAV $i$.

Following Eq. (\ref{eq:dist_update_1})-(\ref{eq:dist_update}), the iteration update for the collaborative VIO problem (\ref{eq:$D^2$VINS_dist}) is,
\begin{align}
    \mathcal{X}_i^{t+1} =& \argmin_{\mathcal{X}_i} f_{cvio_i} (\mathcal{X}_i)   + h(\mathcal{X}_i)
    \notag \\
    &s.t.\  \mathbf{P}_i \in \mathbf{SE}(3),\ \forall \mathbf{P}_i \in \mathcal{T}_i \label{eq:$D^2$VINS_iter_opti} \\
    h(\mathcal{X}_i)& = \frac{1}{2} \sum_{ \mathbf{P}_j \in \mathcal{T}_i}{\Vert \log((\mathbf{P}^t_j)^{-1} \mathbf{P}^k_j) + (\mathbf{\tilde{P}}_j^k)^t \Vert_{\Sigma_{\rho_T}}} \notag\\ 
    & + \rho_\mathcal{L} \sum_{l_i\in \mathcal{L}_i}{\Vert l_i - l_i^t \Vert } \\
    \mathbf{P}_j^{t+1} =& \frac{1}{n_i} \sum^{n_i}_{k=1}(\mathbf{P}_j^k)^{t+1}, \ \forall \mathbf{P}_j \in \mathcal{T}_i \label{eq:$D^2$VINS_averaging} \\
    (\mathbf{\tilde{P}}_j^k)^{t+1} =& (\mathbf{\tilde{P}}_j^k)^{t} + \log((\mathbf{P}_j^{t+1})^{-1} (\mathbf{P}_j^k)^t), \ \forall \mathbf{P}_j \in \mathcal{T}_i ,\label{eq:$D^2$VINS_iter_update}
\end{align}
where $\log(\cdot)$ is the logarithm map of the Lie group SE(3)\cite{blanco2021tutorial}, $(\mathbf{\tilde{P}}_j^k)^t$ is the auxiliary variable, which will be detailed in Sect. \ref{sect:manifold}.
$h(\mathcal{X}_i)$ is the consensus item to ensure the local state $\mathcal{X}_i$ is consistent with the global state $\mathcal{X}$.
Inside $h(\mathcal{X}_i)$, 
the first item $\sum_{ P \in \mathcal{T}_i}{\Vert \log((\mathbf{P}^t_i)^{-1} \mathbf{P}^k_i) + (\tilde{\mathbf{P}^k_i})^t \Vert_{\Sigma_{\rho_T}}}$ is to ensure the poses are consistent among UAVs, and the second item $\rho_\mathcal{L} \sum_{l_i\in \mathcal{L_i}}{\Vert l_i - l_i^t \Vert }$ is to ensure the convergence of the distributed optimization as proved in \cite{zhang2017distributed}.
The full algorithm of the distributed optimization for collaborative VIO is shown in Alg. \ref{alg:$D^2$VINS}.

Subsequent experiments, as detailed in Sect. \ref{sect:eva_$D^2$VINS}, confirm ADMM's suitability for $D^2$VINS.
Importantly, each UAV independently computes Eq. (\ref{eq:$D^2$VINS_averaging}). Due to differing neighbor relationships, UAV $i$ focuses only on averaging poses in $\mathcal{X}_i$. While this involves some computational redundancy, its impact on the overall distributed computing performance is minimal. Additionally, data sharing in Alg. \ref{alg:$D^2$VINS} is limited to UAVs within the near-field state estimation scope.

\begin{algorithm}
\caption{Distributed Optimization for Collaborative VIO on UAV $i$}\label{alg:$D^2$VINS}
\KwIn{$n$ UAVs, $m$ landmarks, $iter$ iterations, $\rho$ and $\rho_T$.}
\KwOut{$n$ UAVs' local states $\mathcal{X}_i$.}
Initialize $\mathcal{X}_i$ and $\mathcal{X}$, $\mathcal{T}_i$, $\mathcal{L}_i$. \\
\For{$t=1$ to $iter$} {
Solve the local optimization problem (\ref{eq:$D^2$VINS_iter_opti}) with $\mathcal{X}_i$ \\
Broadcast the poses $\mathcal{T}_i$ in local state $\mathcal{X}_i$ to all UAVs \\
Update $\mathcal{X}_i$ with by averaging poses among swarm with Eq. (\ref{eq:$D^2$VINS_averaging}). \\
Update the states for auxiliary variables $\{\mathbf{\tilde{P}}_j^k ... \}$ using Eq. (\ref{eq:$D^2$VINS_iter_update}). \\
}
\end{algorithm}

\subsubsection{Optimization on manifold}\label{sect:manifold}
One key problem in SLAM is the optimization on the manifold.
In Alg. \ref{alg:$D^2$VINS}, Problem (\ref{eq:$D^2$VINS_iter_opti}) is solved by Riemannian trust region (RTR) method \cite{boumal2020introduction} to ensure the pose state is on $\mathbf{SE}(3)$ group, i.e., $ \mathbf{P}_i \in \mathbf{SE}(3),\ \forall \mathbf{P}_i \in \mathcal{T}_i$.
In addition, the auxiliary states $\{\mathbf{\tilde{P}}_j^k ... \}$ are defined on  $\mathfrak{se}(3)$. 
We use the logarithm map \cite{blanco2021tutorial} to transform the error $(\mathbf{P}_j^{t+1})^{-1} (\mathbf{P}_j^k)^t$ defined on $\mathbf{SE}(3)$ into $\mathfrak{se}(3)$.
The quaternion averaging in Eq. (\ref{eq:$D^2$VINS_averaging}) is solved by quaternion averaging algorithm \cite{markley2007averaging}, which is well known in the aerospace field.

Our modified ADMM algorithm in Alg. \ref{alg:$D^2$VINS} allows states to be defined on SE(3), enabling us to leverage a wide range of foundational tools from VINS research, such as the Riemannian trust-region method (adopted in \cite{qin2017vins}) and existing marginalization techniques (Sect. \ref{sect:margin}). This approach also helps avoid deviations from the linearization point due to incorrect initialization.
In contrast, for pose estimation problem, ARock shall use variables on $\mathfrak{se}3$ (will be detailed in Sect. \ref{sect:$D^2$PGO}).
When using ARock on Problem \ref{eq:$D^2$VINS_dist}, if the initialization of VIO is suboptimal—which is unavoidable in practice—the results on $\mathfrak{se}3$ may deviate significantly from the linearization point, potentially compromising the accuracy.
We believe above factors positively influence the convergence of our algorithm compared to ARock approach, as verified by our early-stage tests.
What's more, ADMM's robustness in complex, large-scale bundle adjustment scenarios is well-established \cite{eriksson2016consensus, zhang2017distributed}, leading us to select this approach instead of using ARock.

\subsection{Landmark Management}\label{sect:lmanager}
It is essential to partition landmarks to disjoint sets for each UAV (detailed in \ref{sect:dist_ba}), cap the number of landmarks to balance real-time performance demands, while ensuring sufficient measurements for observability in challenging conditions. Moreover, filtering out outlier measurements increases the system's robustness and reduces the likelihood of false matches.
These requirements of landmark partitioning can be formulated as follows, for agent $k$,
\begin{eqnarray} \label{eq:select}
    \max_{\mathcal{L}_k \subseteq \mathcal{L}} & (f_s(\mathcal{L}_k), f_c(\mathcal{L}_k)), \label{eq:mls_multi_obj} \\
    f_s(\mathcal{L}_k) = &  w_l m_l(\mathcal{L}_k) + w_r m_r(\mathcal{L}_k) - w_o o(\mathcal{L}_k), \label{eq:select_cost} \\
    f_c(\mathcal{L}_k) = & \min_{j,t} m(\mathcal{L}_k, \mathcal{K}_j^t) \label{eq:coverage}\\
    \text{s. t. } &|\mathcal{L}_k| \leq	 \tau_l \label{eq:number}\\ 
                  &m_l(\mathcal{L}_k) + m_r(\mathcal{L}_k) \leq \tau_m \label{eq:number_2}\\ 
    & \mathcal{L}_k \subseteq \mathcal{LD}_k \label{eq:discover},
\end{eqnarray}
where $f_s(\mathcal{L}_k)$ is the weighted number of measurements of $\mathcal{L}_k$, $m_l(\mathcal{L}_k)$ and $m_r(\mathcal{L}_k)$ represent the number of measurements of $\mathcal{L}_k$ from local keyframes (keyframes generated by UAV $k$ itself) and remote keyframes, respectively.
$m(\mathcal{L}_k, \mathcal{K}_j^t)$ is the number of measurement of $\mathcal{L}_k$ and keyframe $\mathcal{K}_j^t$, and
$f_c(\mathcal{L}_k)$ is its lower bound. $\max_{\mathcal{L}_k} f_c(\mathcal{L}_k)$ aims to maximize this lower bound.
$o(\mathcal{L}_k)$ is the number of outlier measurements of $\mathcal{L}_k$. 
The weights $w_l$, $w_r$, and $w_o$ dictate the significance of respective measurements. The number of landmarks in $\mathcal{L}_k$ is capped by $\tau_l$, while the maximum number of measurements is restricted by $\tau_m$.  $\mathcal{K}_j^t$ refers to a frame within the sliding window.
$\mathcal{LD}_k$ represents the landmarks first discovered by UAV $k$, so the sets $\{\mathcal{LD}_1, \mathcal{LD}_2, \ldots, \mathcal{LD}_n \}$ partition landmarks $\mathcal{L}$ into $n$ disjoint groups and the condition $\mathcal{L}_k \subseteq \mathcal{LD}_k$ ensures that $\mathcal{L}$ is divided into disjoint sets in Problem (\ref{eq:mls_multi_obj}). 
$D^2$VINS adopts an 'observed-first' principle for this partitioning (in Eq. ({\ref{eq:discover}})), which primarily benefits from single-round communication during complete keyframe broadcasts.
While alternative strategies like graph partitioning techniques \cite{xu2021bdpgo} could potentially enhance optimization, they may introduce extra communication overhead. Therefore, such methods are not implemented in $D^2$VINS.

\begin{algorithm}[t]
    \caption{Multi-Robot Landmark Selection}\label{alg:selection}
    \KwIn{Current UAV $k$, available landmarks $\mathcal{L}$, set of available keyframes $\{\mathcal{F}\}$, a dictionary $\mathcal{O}$ representing the observed landmarks of each frame, with the relationship $\mathcal{O}[\mathcal{F}]\rightarrow\mathcal{L}_f$.}
    \KwOut{Set of selected landmarks $\mathcal{L}_k$}
    $\mathcal{N}_l \leftarrow \emptyset $ \\
    $\mathcal{L}_k \leftarrow \emptyset $ \\
    $\mathcal{O}_a \leftarrow \mathcal{O}$ \\
    \ForEach{$\mathcal{F} \in \{\mathcal{F}\}$ }{
        $\mathcal{N}_l[\mathcal{F}] \leftarrow 0$
    }
    \While{True}{
        \If{$empty(\mathcal{N}_l$)} {
            break
        }
        $\mathcal{F} \leftarrow \argmin_{\mathcal{F} \in \{\mathcal{F}\}}{\mathcal{N}_l[\mathcal{F}]} $ \label{alg:find_f} \\
        \If{$empty(\mathcal{O}_a[\mathcal{F}])$} {
            $\mathcal{N}_l \leftarrow \mathcal{N}_l \setminus \{\mathcal{F}\}$ \\
            continue
        }
        $l \leftarrow \argmax_{l \in \mathcal{O}_a[\mathcal{F}]} cost(k)$ \\
        \ForEach{$track \in l.tracks$ }{
            $\mathcal{N}_l[track.\mathcal{F}] \leftarrow \mathcal{N}_l[track.\mathcal{F}] + 1$ \\
        }
        $\mathcal{L}_k \leftarrow \mathcal{L}_k \cup \{l\}$ \\
        \ForEach{$\mathcal{F} \in \{\mathcal{F}\}$ }{
            $\mathcal{O}_a[\mathcal{F}] \leftarrow \mathcal{O}_a[\mathcal{F}] \setminus \{l\}$ 
        }
        \If{$|\mathcal{L}_k| > \tau_l$ or $m_l(\mathcal{L}_k) + m_r(\mathcal{L}_k) > \tau_m$} {
            break
        }
    }

\end{algorithm}

Problem (\ref{eq:select})  is an extended coverage problem, which involves multi-objective optimization.
To solve this, we introduce a novel multi-robot landmark selection (MLS) algorithm, detailed in Alg. \ref{alg:selection}. Within this algorithm, we use the following cost function to approximate Eq. (\ref{eq:select_cost}):
\begin{equation}\label{eq:cost}
    cost(l) =
\begin{cases}
    w_l m_l(l) + w_r m_r(l) - w_o o(l) & \text{if } l \in \mathcal{LD}_k \\
-1 & \text{otherwise},
\end{cases}
\end{equation}
where $m_l(l)$ and $m_r(l)$ represent the number of measurements of landmark $l$ from local and remote keyframes, respectively, and $o(l)$ is the count of outlier measurements for $l$.

To select landmarks, our algorithm initially sets up dictionary $\mathcal{N}_l$ to track each landmark's current measurement count and dictionary $\mathcal{O}_a$ to record unselected landmarks observed by each frame.
Each iteration begins by identifying the keyframe with the fewest measurements in the sliding window (line \ref{alg:find_f}), aiming to approximate Eq. (\ref{eq:coverage}).
It then selects a landmark $l$ observed by this keyframe that maximizes Eq. (\ref{eq:cost}). Subsequently, the algorithm updates $\mathcal{N}_l$, incrementing $l$'s measurement count for each corresponding keyframe in $l.tracks$. The selected landmark $l$ is added to $\mathcal{L}_k$, and $\mathcal{O}_a$ is updated to exclude $l$ from each keyframe's observed landmarks. The process repeats until the number of selected landmarks surpasses $\tau_l$, the count of selected measurements reaches $\tau_m$, or no more measurements are available. The resulting set of selected landmarks is $\mathcal{L}_k$.

Given the occasional scarcity of common landmarks among robots, landmarks may be reused across different robots in $D^2$VINS, a situation that arises when Eq. (\ref{eq:discover}) is not fully met.
This reuse increases the computational burden and can slightly compromise accuracy, yet it is vital for the system's robustness.
Our experiments have shown that the system can become unstable or even diverge when measurements per frame are few, necessitating a compromise in Alg. \ref{alg:selection} for stability.
In practice, we set $w_l=1$ and $w_r=2$ to encourage the addition of more cross-robot feature points and $w_o=2$ to reduce outliers, identified by large reprojection errors prior to optimization.

\subsection{Sliding Window \& Marginalization}\label{sect:margin}
\subsubsection{Sliding window}
In $D^2$VINS, each UAV independently manages its keyframes within a sliding window, akin to the approach in \cite{qin2017vins}. Upon adding a new frame, if it is not a keyframe, the second newest frame is discarded; otherwise, the oldest keyframe is removed. Following each update of the sliding window, the latest information is shared with other UAVs in the swarm.

\subsubsection{Management of remote keyframes}
Successful multi-robot sparse feature matching, as detailed in Sect. \ref{sect:multi-UAV-match}, results in the inclusion of the remote frame's pose $\tensor*[^{v_k}]{\mathbf{\hat T}}{_j^t}$ into the local state vector $\mathcal{X}_i$ for optimization in Problem (\ref{eq:$D^2$VINS_dist}).
Upon receiving updated sliding window data from other UAVs, $D^2$SLAM removes any remote keyframes from UAV $k$ that are no longer within its local sliding window. This removal, akin to local keyframe deletion, leads to marginalization, which will be discussed further.

\subsubsection{Marginalization} 
When discarding old keyframes, each subproblem $f_{cvio_i}(\mathcal{X}i)$ of Problem (\ref{eq:$D^2$VINS_dist}) is linearized, and a new set of priors $\{ (\mathbf{r}_{p_0}, \mathbf{H}_{p_0}), (\mathbf{r}_{p_1}, \mathbf{H}_{p_1}) \cdots (\mathbf{r}_{p_{N-1}}, \mathbf{H}_{p_{N-1}}) \} $ is computed using the Schur complement to exclude old states.
This process, known as marginalization \cite{qin2017vins, leutenegger2015keyframe}, treats each subproblem independently and is therefore executed distributively.
Since Problem (\ref{eq:$D^2$VINS_dist}) involves states averaged across UAVs (Eq. \ref{eq:$D^2$VINS_averaging}), the marginalization process maintains consistency across UAVs, meaning states are linearized at similar or identical points on different UAVs.

\subsection{Initialization}
$D^2$VINS initialization involves two phases: local keyframe initialization and remote UAV initialization.
For local keyframes, utilizing stereo or multi-camera setups, we employ triangulation for landmark position initialization where measurements have a sufficient baseline (typically 0.05cm) from multiple cameras or due to motion.
Additionally, IMU predictions initialize poses of subsequent local UAV Keyframes.
To enhance stability, priors are added to the unobservable components ($x, y, z, yaw$) of the first keyframe.
In multi-UAV scenarios, PnP RANSAC (or UPnP) is used to initialize the pose of remote keyframes when their coordinate system references differ from the local UAV.

\subsection{Implementation}
$D^2$SLAM solves Problem (\ref{eq:$D^2$VINS_iter_opti}) using Ceres-solver \cite{ceres-solver}, applying a Dogleg strategy \cite{madsen2004methods} and a dense Schur solver. For enforcing pose manifold constraints, we utilize the Local Parameterization feature in Ceres, essentially implementing the RTR (Riemannian Trust-Region) method \cite{boumal2020introduction}.

Alg. \ref{alg:$D^2$VINS}, due to Eq. (\ref{eq:$D^2$VINS_averaging}), operates synchronously, requiring UAVs to wait for others' results each iteration, which can be less efficient in poor communication environments. However, we've found that relaxing the synchronization condition of Eq. (\ref{eq:$D^2$VINS_averaging}) — using the most recently received state averages without waiting — also produces results close to the synchronous approach. We label this method as asynchronous $D^2$VINS, and a comparative analysis with its synchronous counterpart will be presented in Sect. \ref{sect:eva_$D^2$VINS}.

\section{Far-field state estimation: $D^2$PGO}\label{sect:$D^2$PGO}
Far-field estimation in $D^2$SLAM encompasses both global trajectory consistency ($D^2$PGO) and real-time global odometry. We model the global trajectory estimation using a pose graph approach \cite{grisetti2010tutorial}, where keyframes are graph nodes and edges represent relative poses between them. This pose graph is formulated as a factor graph, treating measurements as Gaussian-distributed, thus constituting the classic pose graph optimization problem \cite{dellaert2017factor}. $D^2$PGO employs a two-stage optimization process to decentralize and distribute the pose graph optimization across multiple UAVs.
The pose graph optimization problem is,
\begin{equation}\label{eq:PGO}
    \begin{aligned}
        \min_{\mathbf{x}_i} &\sum_{ (i_{t_0}, j_{t_1}) \in \mathcal{E}} \left\Vert { }^{v_k} \mathbf{\hat p}_{j}^{t_1} - { }^{v_k} \mathbf{\hat p}_{i}^{t_0} - { }^{v_k} \mathbf{\hat R}_i^{t_0} {\mathbf{z}_\mathbf{p}}_{i\rightarrow j}^{t_0 \rightarrow t_1} \right\Vert^2_{\Sigma_t} \\
        & + \left\Vert { }^{v_k}\mathbf{\hat R}_{j}^{t_1} - { }^{v_k}\mathbf{\hat R}_{i}^{t_0} {\mathbf{z}_\mathbf{R}}_{i\rightarrow j}^{t_0 \rightarrow t_1} \right\Vert^2_{\Sigma_R}, \\
          &s.t.\ { }^{v_k}\mathbf{\hat R}_i^{t}\in SO(3),
    \end{aligned}
\end{equation}
where $\mathcal{E}$ is the set of all edge, including loop closure edge (generated by loop closure detection) and ego-motion edge (generated by $D^2$VINS),
${\mathbf{z}_\mathbf{p}}_{i\rightarrow j}^{t_0 \rightarrow t_1}, {\mathbf{z}_\mathbf{R}}_{i\rightarrow j}^{t_0 \rightarrow t_1}$ are the rotation part and translation part of the relative pose of the edge $(i_{t_0}, j_{t_1})$, respectively.
${ }^{v_k}\mathbf{R}_i^{t}$ is the rotation part of $ { }^{v_k}\mathbf{\hat T}_i^{t}$, and ${ }^{v_k}\mathbf{p}_i^{t}$ is the translation part of $ { }^{v_k}\mathbf{\hat T}_i^{t}$, $\mathbf{x}$ is the full state of the pose graph:
\begin{equation}
    \begin{aligned}
        \mathcal{X} = \Big[& ^{v_k}\mathbf{\hat T}_{0}^{t_0},  ^{v_k}\mathbf{\hat T}_{0}^{t_1}, \cdots ^{v_k}\mathbf{\hat T}_{0}^{t_{M - 1}}, \cdots\ ^{v_k}\mathbf{\hat T}_{N-1}^{t_0}, \\ 
        &\ ^{v_k}\mathbf{\hat T}_{N-1}^{t_1},\cdots ^{v_k}\mathbf{\hat T}_{N-1}^{t_{M - 1}}\Big],
    \end{aligned}
\end{equation}
where $N$ is the number of UAVs, $M$ is the number of keyframes of each UAV.
        
By reformulating the pose graph optimization problem (\ref{eq:PGO}) into a distributed framework (\ref{eq:dist_problem}), the equation becomes:
\begin{equation}\label{eq:DPGOSLAM}
    \begin{aligned}
        &\min_{\mathbf{x}_i} \sum_{i=0}^{N-1} f_{pgo_i}(\mathbf{x}_i)&& \\
        f_{pgo_i}(\mathbf{x}_i)& = \sum_{ (i_{t_0}, j_{t_1}) \in \mathcal{E}^i} \left\Vert { }^{v_k} \mathbf{\hat p}_{j}^{t_1} - { }^{v_k} \mathbf{\hat p}_{i}^{t_0} - { }^{v_k} \mathbf{\hat R}_i^{t_0} {\mathbf{z}_\mathbf{p}}_{i\rightarrow j}^{t_0 \rightarrow t_1} \right\Vert^2_{\Sigma_p}  \\ + & \left\Vert { }^{v_k}\mathbf{\hat R}_{j}^{t_1} - { }^{v_k}\mathbf{\hat R}_{i}^{t_0} {\mathbf{z}_\mathbf{R}}_{i\rightarrow j}^{t_0 \rightarrow t_1} \right\Vert^2_{\Sigma_R}, &&\\
          &s.t.\ { }^{v_k}\mathbf{\hat R}_i^{t} = \mathbf{E}^\mathbf{R}_{i_t} \mathbf{z}, { }^{v_k}\mathbf{\hat p}_i^t = \mathbf{E}^p_{i_t} \mathbf{z}, 
          { }^{v_k}\mathbf{R}_i^{t}\in SO(3),&
    \end{aligned}
\end{equation}
where $x_i$ is the state of the UAV $i$, $ \mathbf{E}^\mathbf{R}_{i_t} \mathbf{z}, \mathbf{E}^p_{i_t}$ project global state $z$ to local states, $\mathcal{E}^i$ is the set of edges of the UAV $i$.

$D^2$PGO's initial stage employs rotation initialization for the pose graph to circumvent local minima. The second stage focuses on refining the pose graph optimization. Beyond optimization, an outlier rejection module is implemented to filter anomalies in the pose graph problem (\ref{eq:PGO}). The final step involves merging the pose graph optimization results with the VIO results to establish each UAV's globally consistent real-time odometry:
\begin{equation}\label{eq:forward_pro}
    \tensor*[^{v_k}]{\mathbf{\hat T}}{_i^{t_1}} = \tensor*[^{v_k}]{\mathbf{\hat T}}{_i^{t}}(\tensor*[]{\mathbf{\tilde T}}{_i^{t}})^{-1} \tensor*[]{\mathbf{\tilde T}}{_i^{t_1}},
\end{equation}
where $\tensor*[^{v_k}]{\mathbf{\hat T}}{_i^{t}}$ denotes the latest keyframe pose of UAV $i$ in $D^2$PGO, $\tensor*[]{\mathbf{\tilde T}}{_i^{t}}$ represents the odometry estimation from $D^2$VINS for this keyframe, and $\tensor*[]{\mathbf{\tilde T}}{_i^{t_1}}$ is the real-time odometry estimation with $D^2$VINS.

\subsection{ARock for Distributed Optimization}
To solve the PGO, we employ ARock \cite{peng2016arock}, an asynchronous distributed optimization algorithm. ARock iteratively updates the distributed optimization problem (\ref{eq:dist_problem}) as described in \cite{peng2016arock}:
\begin{align}
    \mathbf{\hat x}_i^t = & \argmin_{\mathbf{x}_i} f_i(\mathbf{x_i}) + \mathbf{x_i} \sum_{r\in \mathcal{R}(i)} z^t_{ri,r} + \frac{\gamma}{2} \vert\mathcal{R}(i)\vert \cdot \Vert \mathbf{x_i} \Vert^2 \\
    z^{t+1}_{ri,i} = & z_{ri,i}^t  - \eta_k ((z^{t}_{ri,i} + z_{ri,r})/2 + \gamma \mathbf{\hat x}_i^t)  \ \forall r\in \mathcal{R}(i),
\end{align}
where $\mathcal{R}(i)$ represents the set of neighbors for agent $i$, $(\cdot)^t$ denotes the state in the $t$-th iteration, and $\eta_k$ and $\gamma$ are two parameters.
Dual variables $z_{ij,i}$ and $z_{ij,j}$ for agents $i$ and $j$ are updated independently by each, and $z_{ij,i} = z_{ji,i}, z_{ij,j} = z_{ji,j}$.
We modify the update rule by substituting dual variables with $y_{ij,i} = - z_{ij,i}/\gamma$ and $y_{ij,j} = - z_{ij,j}/\gamma$ to obtain:
\begin{align}
    \mathbf{\hat x}_i^t = & \argmin_{\mathbf{x}_i} f_i(\mathbf{x_i}) + \sum_{r\in \mathcal{R}(i)} \frac{\gamma}{2} \Vert \mathbf{x_i} - y^t_{ri,r} \Vert^2  \label{eq:arock_opti} \\
    y^{t+1}_{ri,i} = & y_{ri,i}^t  - \eta_k ((y^t_{ri,i} + y_{ri,r})/2 - \mathbf{\hat x}_i^t) \ \forall r\in \mathcal{R}(i)  \label{eq:arock_update}
\end{align}
The complete ARock algorithm for decentralized optimization is outlined in Alg. \ref{alg:arock}. Unlike the synchronous Alg. \ref{alg:$D^2$VINS}, ARock operates asynchronously, utilizing the latest received dual variables from remote agents without the need for synchronous execution as per Eq. (\ref{eq:arock_opti})-(\ref{eq:arock_update}), making it less sensitive to communication delays. Additionally, ARock is proven to have linear convergence \cite{peng2016arock}.

\begin{algorithm}[t]
\caption{ARock for Distributed Optimization}\label{alg:arock}
\KwIn{$n$ UAVs, $m$ landmarks, $iter$ iterations.}
Initialize $y_{ij,i}^0, y_{ij,j}^0$ for all $i,j$ to $0$. \\
\For{$t=1$ to $iter$} {
Solve the local optimization problem (\ref{eq:arock_opti}) \\
Update the dual states $y^t_{ri,i}$ with Eq. (\ref{eq:arock_update}). \\
Broadcast the dual state $y^t_{ri,i}$ to all UAVs \\
\If{Convergence} {
    Break
}
}
\end{algorithm}
\subsection{Asynchronous Distributed Rotation Initialization}
Nonlinearities in the pose graph problem primarily arise from rotations.
With proper rotation initialization, the problem closely resembles a linear least squares issue, simplifying its solution.
This technique of rotation initialization is known to effectively avoid local minima in pose graph optimization \cite{carlone2015initialization}.
The goal of this initialization is to find an exact or approximate solution for the rotational component of Problem (\ref{eq:PGO}), a task akin to solving the rotation averaging problem \cite{hartley2013rotation}:
\begin{equation}\label{eq:rot_init}
    \begin{aligned}
        \min_{\mathbf{x}_i} & \sum_{ (i_{t_0}, j_{t_1}) \in \mathcal{E}^i}{\left\Vert  { }^{v_k}\mathbf{\hat R}_{j}^{t_1} - { }^{v_k}\mathbf{\hat R}_{i}^{t_0} {\mathbf{z}_\mathbf{R}}_{i\rightarrow j}^{t_0 \rightarrow t_1} \right\Vert^2_{\Sigma_R}}\\
          &s.t.\ { }^{v_k}\mathbf{\hat R}_i^{t}\in SO(3).
    \end{aligned}
\end{equation}
Chordal relaxation \cite{martinec2007robust} is an effective algorithm for solving the rotation initialization problem. It starts by relaxing the $\mathbf{SO}(3)$ constraint in (\ref{eq:rot_init}):
\begin{equation}\label{eq:chord}
    \begin{aligned}
        \min_{\mathbf{x}_i} \sum_{ (i_{t_0}, j_{t_1}) \in \mathcal{E}}&{ \left\Vert { }^{v_k}\mathbf{\overline R}_{j}^{t_1} - { }^{v_k}\mathbf{\overline R}_{i}^{t_0} {\mathbf{z}_\mathbf{R}}_{i\rightarrow j}^{t_0 \rightarrow t_1} \right\Vert^2_{\Sigma_R}}  \\ 
        & +  \sum_{{\overline{R}}_{i}^{t} \in \mathcal{R}_i}{\Vert { }^{v_k}(\mathbf{\overline{R}}_{i}^{t})_3 - \mathbf{v}_i \Vert}_{\Sigma_g},
    \end{aligned}
\end{equation}
where $\mathcal{R}_i$ denotes all rotations in $\mathbf{x}_i$, and $\mathbf{\overline R}_{i}^{t}$ is a 3x3 matrix representing the rotation of UAV $i$ at time $t$. The term ${\Vert { }^{v_k}(\mathbf{\overline{R}}_{i}^{t})_3 - \mathbf{v}_i \Vert}_{\Sigma_g}$ is the vertical prior, added because the roll-pitch angle from VIO is observable and can be used as a prior, enhancing initialization \cite{carlone2015initialization}. Here, $ { }^{v_k}(\mathbf{\overline{R}}_{i}^{t})_3$ is the third row of $\mathbf{\overline{R}}_{i}^{t}$, $\mathbf{v}_i = (\mathbf{\tilde R}_{i}^{t})^T \mathbf{g}$ with $g=[0, 0, 1]^T$, and $\mathbf{\tilde R}_{i}^{t}$ is the odometry output rotation of UAV $i$ at time $t$. Problem (\ref{eq:chord}) is a linear least squares problem that can be efficiently solved with a linear solver.
Then, we can recover the rotation matrix by
\begin{equation}\label{eq:recover}
    \begin{aligned}
    { }^{v_k}\mathbf{\hat R}_{i}^{t} = &argmin_{\mathbf{\hat R}} \Vert \mathbf{\hat R} - { }^{v_k}\mathbf{\overline R}_{i}^{t} \Vert^2_F, \\
    &s.t.\ \hat R \in \mathbf{SO}(3)
\end{aligned}
\end{equation}
where $\Vert \cdot \Vert_F$ is the Frobenius norm. 
Problem (\ref{eq:recover}) has a closed form soluting with SVD decomposition \cite{hartley2013rotation}:
\begin{equation}\label{eq:recover2}
    { }^{v_k}\mathbf{\breve R}_{i}^{t} = \mathbf{S} diag([1, 1, det(\mathbf{SV^T})]) \mathbf{V^T},
\end{equation}
where ${ }^{v_k}\mathbf{\breve R}_{i}^{t}\in SO(3)$ is the initial result of ${ }^{v_k}\mathbf{\hat R}_{i}^{t}$, $\mathbf{S D V^T}$ is the SVD decomposition of $\mathbf{\overline R}_{i}^{t}$.

In $D^2$PGO, we solve Problem (\ref{eq:chord}) distributively, redefining it as:
\begin{equation}\label{eq:chord_dist}
    \begin{aligned}
        \min_{\mathbf{x}_i} \sum_{i=0}^{N-1}&\Bigg\{\sum_{ (i_{t_0}, j_{t_1}) \in \mathcal{E}_i}{ \left\Vert { }^{v_k}\mathbf{\overline R}_{j}^{t_1} - { }^{v_k}\mathbf{\overline R}_{i}^{t_0} {\mathbf{z}_\mathbf{R}}_{i\rightarrow j}^{t_0 \rightarrow t_1} \right\Vert^2_{\Sigma_R}} \\ 
        & + \sum_{{\overline{R}}_{i}^{t} \in \mathbf{x}_i}{\Vert { }^{v_k}(\mathbf{\overline{R}}_{i}^{t})_3 - \mathbf{v}_i \Vert_{\Sigma_g}}\Bigg\},
    \end{aligned}
\end{equation}
Utilizing Alg. \ref{alg:arock}, we address problem (\ref{eq:chord_dist}) through our asynchronous distributed rotation initialization algorithm, detailed in Alg. \ref{alg:dist_rot_init}.
The local optimization problem within Alg. \ref{alg:dist_rot_init} (converting Eq. (\ref{eq:chord_dist}) into Eq. (\ref{eq:arock_opti})) is a linear least squares problem, efficiently solvable via Cholesky factorization \cite{nocedal1999numerical}. Iteration termination occurs when changes in the optimized state become negligible, as indicated in Alg. \ref{alg:dist_rot_init} line \ref{alg:stop_rot_init}, with $(\mathbf{\overline{R}})^t$ representing the post-$t$-iteration state.
Additionally, to avoid premature termination (e.g., due to initial data absence), a minimum number of iterations is also set.

\begin{algorithm}[t]
    \caption{Asynchronous Distributed Rotation Initialization for UAV $i$}\label{alg:dist_rot_init}
    \KwIn{$\epsilon$ for termination, $max\_iter$ iterations, $min\_iter$ to avoid early exiting.}
    Initialize $y_{ij,i}^0, y_{ij,j}^0$ for all $i,j$ to $0$. \\
    \For{$t=1$ to $max\_iter$} {
    Solve the local optimization problem (\ref{eq:arock_opti}) with $f_i$ defined from (\ref{eq:chord_dist}) \\
    Update the dual states $y^t_{ir,i}$ with Eq. (\ref{eq:arock_update}). \\
    Broadcast the dual state $y^t_{ir,i}$ to neighbors\\
    \If{$\frac{1}{\vert\mathcal{R}_i\vert}\sum_{\mathbf{\overline{R}} \in \mathcal{R}_i }{{\Vert(\mathbf{\overline{R}})^t - (\mathbf{\overline{R}})^{t-1}\Vert_F}/{\Vert(\mathbf{\overline{R}})^t\Vert_F } < \epsilon}$ 
    \textbf{and} $iter > min\_iter$ \label{alg:stop_rot_init} }
    {
        break
    }
    }
    \For{$\overline{R} \in \mathcal{R}_i$} {
        Recover the rotation ${ }^{v_k}\mathbf{\hat R}_{i}^{t}$ with Eq. (\ref{eq:recover2}) \\
    }
\end{algorithm}

\begin{algorithm}[t]
    \caption{Asynchronous Distributed Pose Graph Optimization for UAV $i$}\label{alg:dist_pgo}
    \While{Not shutdown}{
        \If{New data received} {
            Adding new states or cost functions to optimizer.
        }
        Solve the local optimization problem (\ref{eq:arock_opti}) with $f_i$ defined from (\ref{eq:dpgo_perturb}) \\
        Update the dual states $y^t_{ir,i}$ with Eq. (\ref{eq:arock_update}). \\
        Broadcast the dual state $y^t_{ir,i}$ to neighbors \\
        Recover the poses with Eq. (\ref{eq:recover_pose}) \\
        Sleep for a while
    }
\end{algorithm}

\subsection{Asynchronous Distributed Pose Graph Optimization}\label{sect:dist_pgo}

After completing the initialization, we build the perturbation problem of (\ref{eq:DPGOSLAM}) with the initialization rotation, 
\begin{equation} \label{eq:dpgo_perturb}
    \begin{aligned}
        &\min_{\mathbf{x}_i^*} \sum_{i=0}^{N-1} f_{pgo_i}^*(\mathbf{x}_i^*)&& \\
        f_{pgo_i}^*(\mathbf{x}_i^*)& = \sum_{ (i_{t_0}, j_{t_1}) \in \mathcal{E}^i} \big\Vert { }^{v_k} \mathbf{\hat p}_{j}^{t_1} - { }^{v_k} \mathbf{\hat p}_{i}^{t_0} - 
        && \\
        & \boxplus({ }^{v_k} \mathbf{\breve R}_i^{t_0}, { }^{v_k} \mathbf{\delta \theta}_i^{t})  {\mathbf{z}_\mathbf{p}}_{i\rightarrow j}^{t_0 \rightarrow t_1} \big\Vert^2_{\Sigma_p} + & \\
        & \left\Vert \boxplus({{ }^{v_k}\mathbf{\breve R}_{j}^{t_1}}, { }^{v_k}{\delta \theta}_{j}^{t_1}) - \boxplus({ }^{v_k} \mathbf{\breve R}_i^{t_0}, { }^{v_k} \mathbf{\delta \theta}_i^{t})  {\mathbf{z}_\mathbf{R}}_{i\rightarrow j}^{t_0 \rightarrow t_1} \right\Vert^2_{\Sigma_R},\\
        &s.t.\ { }^{v_k}\mathbf{\delta \theta}_i^{t} = \mathbf{E}^\mathbf{R}_{i_t} \mathbf{z}^*, { }^{v_k}\mathbf{\hat p}_i^t = \mathbf{E}^p_{i_t} \mathbf{z}^*,
    \end{aligned}
\end{equation}
where
$({ }^{v_k} \mathbf{\hat p}_{i}^{t_0}, { }^{v_k} \mathbf{\delta \theta}_i^{t_0})$ is the perturbation state of ${ }^{v_k} \mathbf{T}_i^{t_0}$, where ${ }^{v_k} \mathbf{\hat p}_{i}^{t_0}$ is the position of  ${ }^{v_k} \mathbf{T}_i^{t_0}$ and ${ }^{v_k} \mathbf{\delta \theta}_i^{t_0} \in \mathfrak{so}3$ is the perturbation state of the rotation.
$x_i^*$ is the the perturbation state of $\mathbf{x}^i$, and $\mathbf{z}^*$ is the global perturbation state,
$\boxplus(\mathbf{R}, \mathbf{\delta \theta}) = \mathbf{R} Exp(\mathbf{\delta \theta}_{i}^{t_0})$ retracts $\mathfrak{so}3$ vector to $\mathbf{SO}(3)$ at $\mathbf{R}$\cite{boumal2020introduction},
where $Exp(\cdot)$ maps $\mathfrak{so}3$ to $SO(3)$ \cite{sola2017quaternion}.
Contrasting with the two-step PGO approach in Eq (10) of \cite{choudhary2017distributed}, our Problem (\ref{eq:dpgo_perturb}) is nonlinear and circumvents the accuracy issues associated with the linearization used in \cite{choudhary2017distributed}.

Our algorithm employs Alg. \ref{alg:arock} to address (\ref{eq:dpgo_perturb}), ensuring asynchronicity. Transforming $f_{pgo_i}^*(\mathbf{x}_i^*)$ per (\ref{eq:arock_opti}) yields a nonlinear least squares problem, solvable via the Levenberg-Marquardt algorithm \cite{more1978levenberg}. Convergence is determined by observing changes in the cost function of $f_{pgo_i}^*(\mathbf{x}_i^*)$. As in Alg. \ref{alg:dist_rot_init}, a minimum iteration count is set to avoid premature termination.
After convergence, we recover the global pose with
\begin{equation} \label{eq:recover_pose}
        { }^{v_k}\mathbf{\hat T}_{i}^{t}{ }^{v_k} = 
        \begin{bmatrix}
            \boxplus({ }^{v_k} \mathbf{\breve R}_i^{t_0}, { }^{v_k} \mathbf{\delta \theta}_i^{t}) &
            \mathbf{\hat p}_{i}^{t} \\
            0 & 1
        \end{bmatrix}
\end{equation}
Our asynchronous distributed pose graph optimization (ARockPGO) is outlined in Alg. \ref{alg:dist_pgo}. Unlike Alg. \ref{alg:dist_rot_init}, ARockPGO operates continuously throughout UAV flight, updating at a fixed frequency (typically 1Hz) and incrementally integrating new information. This approach balances convergence speed against communication load.

\subsection{Convergence}
ARock demonstrates strong convergence performance, suggesting it can potentially reach optimality in the initial stage of linear least squares problems. The effectiveness of rotation initialization in preventing convergence to incorrect local optima \cite{carlone2015initialization} enhances $D^2$PGO's ability to achieve or closely approximate the global optimum.

\subsection{Outlier Rejection}
Currently, the mainstream outlier rejection methods for multi-robot PGO include Pairwise Consistent Measurement Set Maximization (PCM) \cite{mangelson2018pairwise, lajoie2020door, xu2022omni} and Graduated Non-Convexity (GNC) \cite{yang2020graduated}. 
PCM, independent of back-end optimization, is efficient but less effective with large pose graphs.
Conversely, GNC is integrated into back-end optimization, offering higher efficiency but operating synchronously.
In $D^2$PGO, we adopt a distributed version of PCM as used in \cite{xu2022omni, lajoie2020door}, mitigating PCM's efficiency issues at scale. We believe this approach provides adequate performance for PGO that is not real-time critical.

\subsection{Implementation}
The initial stage of $D^2$PGO is designed to ensure accurate trajectory estimation even in highly noisy environments. However, in real-world applications, this stage is often unnecessary, as VIO's yaw measurements are typically precise, allowing for its omission in practice.

$D^2$PGO is implemented in C++, utilizing Eigen\cite{eigenweb}'s LLT algorithm for the linear least squares calculation in rotation initialization.
For the second stage, we employ the Levenberg-Marquardt algorithm from Ceres-solver \cite{ceres-solver}.

\section{System Setup}

\begin{figure}[h]
    \centering
        \centering
        \includegraphics[width=0.7\linewidth]{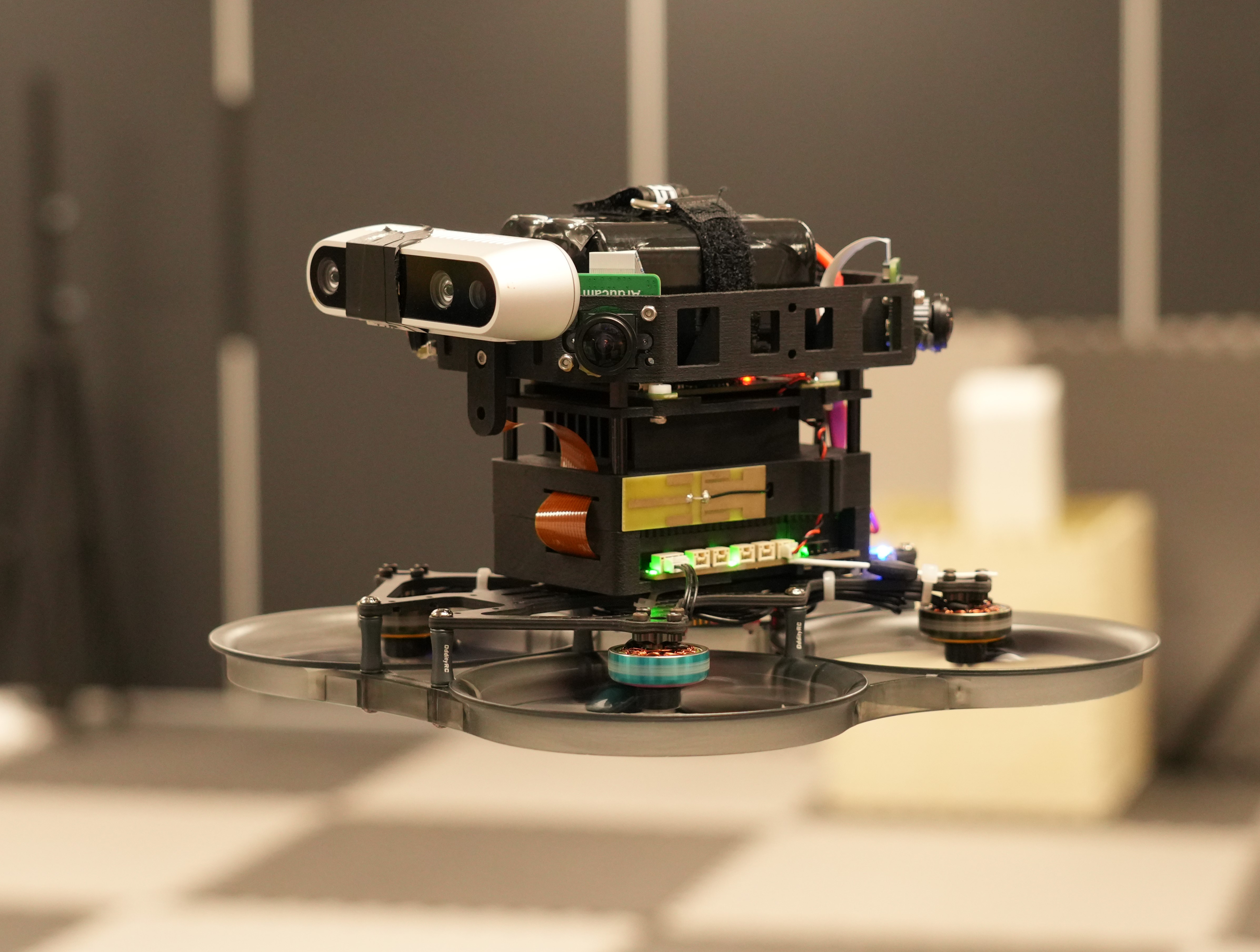}
    \caption{\small{Our aerial platform equipped with quad fisheye cameras, an onboard computer, an all-in-one flight controller, an additional Intel RealSense d435i camera, and a 6s 18650 battery.}}\label{fig:UAV2022}
\end{figure}

\begin{table*}[ht!]
    \caption{The table displays the datasets used for validation. The numbers added as suffix to the datasets correspond to the number of robots involved.}
    \label{tab:datasets}
    \centering
    \begin{tabular}{c|c|c|c|c|c}
    \hline\hline
    Dataset         & \begin{tabular}[c]{@{}c@{}}Robot\\ Number\end{tabular} & Sensors                                                                       & Movement                                                                 & Groundtruth         & Source                                                   \\ \hline\hline
    TUM ROOM 2-5    & 2 to 5                                                 & \multirow{2}{*}{\begin{tabular}[c]{@{}c@{}}Stereo Camera\\ IMU\end{tabular}} & \multirow{3}{*}{HandHold}                                                 & Yes                  & \cite{schubert2018tum}                                   \\ \cline{1-2} \cline{5-5}
    TUM Corr 2-5    & 2 to 5                                                 &                                                                              &                                                                           & Partial              &                                                          \\ \cline{1-3} \cline{5-6} 
    HKUST RI 3          & 3                                                  & \begin{tabular}[c]{@{}c@{}}Stereo Camera\\ Depth Camera\\ IMU\end{tabular}   &                                                                           & No                   & \multirow{4}{*}{Customized}                              \\ \cline{1-5}
    Omni 2-5        & 2 to 5                                                 & \multirow{5}{*}{\begin{tabular}[c]{@{}c@{}}Quad fisheye cameras\\ IMU\end{tabular}}   & Flight                                                                    & \multirow{3}{*}{Yes} &                                                          \\ \cline{1-2} \cline{4-4}
    OmniLongNoYaw 5 & 5                                                      &                                                                              & \begin{tabular}[c]{@{}c@{}}Flight with Yaw\\ Angle Fixed\end{tabular}     &                      &                                                          \\ \cline{1-2} \cline{4-4}
    OmniLongYaw 5   & 5                                                      &                                                                              & \begin{tabular}[c]{@{}c@{}}Flight with Yaw\\  Angle Moving\end{tabular}   &                      &                                                          \\ \hline\hline
    \end{tabular}
\end{table*}

Our experimental aerial swarm is based on custom-modified aerial robots, as depicted in Fig. \ref{fig:UAV2022}. Each UAV is a modified 3.5-inch cinewhoop FPV UAV, equipped with a quartet of fisheye cameras providing 360-degree horizontal visibility. They are powered by NVIDIA Xavier NX onboard computers, boasting robust GPU capabilities. To minimize takeoff weight, we use a lightweight, integrated flight controller with four electronic speed controls (ESCs) and customized PX4 \cite{meier2015px4} firmware. The UAV's takeoff weight is approximately 623g with a 22.8v 1550mah battery, offering a maximum endurance of about 10 minutes, which suffices for our tests. Optionally, an Intel RealSense d435i camera can be mounted on the UAV.

$D^2$SLAM is implemented in C++ and utilizes ONNX Runtime \cite{onnxruntime} for CNN inference. We activate TensorRT for enhanced performance, employing the int8 mode on the onboard computer to optimize speed. Additionally, we develop TaichiSLAM, a GPU-accelerated mapping module, serving as the dense mapping backend for later experiments. TaichiSLAM, crafted in the high-performance Taichi language \cite{hu2019taichi}, integrates various algorithms like OctoMap\cite{wurm2010octomap}, TSDF \& ESDF \cite{oleynikova2017voxblox}, and submap fusion \cite{reijgwart2019voxgraph} to create globally consistent maps. However, as dense mapping algorithms are not the primary focus of $D^2$SLAM, they are not elaborated in detail.

\section{Experiments}
\subsection{Datasets \& Evaluation Setup}

In the evaluation presented in this section, we test $D^2$SLAM using various publicly available and customized datasets. 
These datasets use either stereo cameras or quad cameras as input and partly used a motion capture system to capture ground truth data, as shown in Tab. \ref{tab:datasets}. 
To simulate the scenario of multiple robots operating simultaneously, we align multiple segments of data into a single multi-robot dataset.

Details on the evaluation metrics are available in \cite{xu2022omni}. Notably, the term \textit{Relative Error (RE)} in this context refers to the accuracy of relative state estimation among UAVs. 
In the following tables, position measurements are in meters and rotation in degrees.

In real-time scenarios, we do not require distributed optimization to fully converge, similar to some single-robot SLAM systems like VINS-Mono \cite{qin2017vins}, which typically limits backend optimization to just 8 iterations. 
Synchronous $D^2$VINS performs 4 iterations per solution of Eqs. (\ref{eq:$D^2$VINS_iter_opti}) - (\ref{eq:$D^2$VINS_iter_update}), while asynchronous $D^2$VINS requires only a single run of these equations per optimization.
Given that $D^2$VINS continuously integrates new data and commences optimization from the previous results, with most of the states from the last optimization still included in $\mathcal{X}_i$, Eqs. (\ref{eq:$D^2$VINS_iter_opti}) - (\ref{eq:$D^2$VINS_iter_update})  effectively experience multiple iterations over time in the asynchronous $D^2$VINS framework.
$D^2$PGO's handling in real-time systems is similar to asynchronous $D^2$VINS.

\subsection{Evaluation of $D^2$VINS} \label{sect:eva_$D^2$VINS}
Our initial tests focused on $D^2$VINS to assess its near-field state estimation performance. These tests did not rely on ground truth for initializing $D^2$VINS.
We also compared $D^2$VINS with its single-robot variant, referred to as $D^2$VINS (single), and with the state-of-the-art VIO approach, VINS-Mono\cite{qin2017vins}.
For alignment, VINS-Mono and $D^2$VINS (single) are synchronized using the initial pose from ground truth, mirroring the known UAV departure points setup used in \cite{lusk2020distributed}.

\begin{table*}[h]
    \centering
    \caption{\small{The table displays the statistical comparison of $D^2$VINS against VINS-Mono and $D^2$VINS (single)} on the TUM VI room dataset, where ATE and RE are defined as \cite{xu2022omni}. These datasets exhibit good common FoV during the initial 30 seconds, hence the results are separately presented for this duration, marked as (30s).}
    \label{tab:tum}
    \begin{tabular}{c|c|c|c|c|c|c}
    \hline\hline
    Dataset                             & Avg. Traj. Len.        & Method              & $ATE_{pos}$    & $ATE_{rot}$   & $RE_{pos}$     & $RE_{rot}$    \\ \hline\hline
    \multirow{3}{*}{TUM ROOM 3 (30s)}   & \multirow{3}{*}{18.3} & $D^2$VINS           & \textbf{0.030} & 0.99          & \textbf{0.025} & 0.77          \\ \cline{3-7} 
                                        &                        & $D^2$VINS (async)   & 0.035          & 1.63          & \textbf{0.025} & 0.74 \\ \cline{3-7} 
                                        &                        & $D^2$VINS (single)  & 0.054          & 2.44          & 0.055          & 1.72          \\ \cline{3-7} 
                                        &                        & VINS-Mono           & 0.055          & \textbf{0.66} & 0.072          & \textbf{0.52} \\ \hline\hline
    \multirow{3}{*}{TUM ROOM 3}         & \multirow{3}{*}{141.6} & $D^2$VINS      & 0.128          & 2.40 & 0.119          & 2.95     \\ \cline{3-7} 
                                        &                        &  $D^2$VINS (async)  & \textbf{0.111} & 2.63          & \textbf{0.098} & 3.08          \\ \cline{3-7} 
                                        &                        &  $D^2$VINS (single) & 0.181          & 3.16          & 0.161          & 2.91          \\ \cline{3-7} 
                                        &                        & VINS-Mono           & 0.139          & \textbf{1.49}          & 0.212          & \textbf{2.11} \\ \hline\hline
    \multirow{3}{*}{TUM ROOM 4 (30s)}    & \multirow{3}{*}{19.1}  & $D^2$VINS      & 0.057          & 2.20          & \textbf{0.020} & 0.74    \\ \cline{3-7} 
                                    &                        &  $D^2$VINS (async)  & 0.124          & 1.73          & 0.038          & 0.78          \\ \cline{3-7} 
                                    &                        &  $D^2$VINS (single) & 0.067          & 3.95          & 0.064          & 1.86          \\ \cline{3-7} 
                                    &                        & VINS-Mono           & \textbf{0.056} & \textbf{0.67} & 0.063          & \textbf{0.52}          \\ \hline\hline
    \multirow{3}{*}{TUM ROOM 4}          & \multirow{3}{*}{100.9} & $D^2$VINS      & 0.115          & 2.44          & \textbf{0.080} & 1.72          \\ \cline{3-7} 
                                    &                        &  $D^2$VINS (async)  & \textbf{0.105} & 2.18 & 0.082          & 1.85          \\ \cline{3-7} 
                                    &                        &  $D^2$VINS (single) & 0.178          & 4.40          & 0.144          & 2.88          \\ \cline{3-7} 
                                    &                        & VINS-Mono           & 0.125          & \textbf{1.37} & 0.156          & \textbf{1.55} \\ \hline\hline
    \multirow{3}{*}{TUM ROOM 5 (30s)}   & \multirow{3}{*}{19.2}  & $D^2$VINS      & 0.079          & 1.35          & 0.039          & 0.73 \\ \cline{3-7} 
                                        &                        &  $D^2$VINS (async)  & 0.108          & 4.50          & \textbf{0.037} & 0.89          \\ \cline{3-7} 
                                        &                        &  $D^2$VINS (single) & 0.068          & 3.81          & 0.090          & 1.73          \\ \cline{3-7} 
                                        &                        & VINS-Mono           & \textbf{0.053} & \textbf{0.69} & 0.058          & \textbf{0.65} \\ \hline\hline
    \multirow{3}{*}{TUM ROOM 5}         & \multirow{3}{*}{103.5} & $D^2$VINS      & \textbf{0.090} & 1.85          & \textbf{0.073} & \textbf{1.57} \\ \cline{3-7} 
                                        &                        &  $D^2$VINS (async)  & 0.134          & 4.05          & 0.091          & 2.10          \\ \cline{3-7} 
                                        &                        &  $D^2$VINS (single) & 0.194          & 4.26          & 0.183          & 3.28          \\ \cline{3-7} 
                                        &                        & VINS-Mono           & 0.164          & \textbf{1.62} & 0.209          & 2.28          \\ \hline\hline
    \end{tabular}
\end{table*}

\begin{table*}[h]
    \centering
    \caption{\small{The table presents the statistical comparison of $D^2$VINS with $D^2$VINS (single) and VINS-Mono \cite{qin2017vins} using Omni datasets. It also includes the average distance (Avg. Dis.) between UAVs.}}
    \label{tab:omni}
    \begin{tabular}{c|c|c|c|c|c|c|c}
    \hline\hline
    Dataset                                 & Avg. Traj. Len.       & Avg. Dis.             & Method            & $ATE_{pos}$    & $ATE_{rot}$    & $RE_{pos}$     & $RE_{rot}$    \\ \hline\hline
    \multirow{2}{*}{Omni 2}                 & \multirow{3}{*}{49.1} & \multirow{3}{*}{0.68} & $D^2$VINS         & \textbf{0.069} & \textbf{0.85}  & \textbf{0.075} & 0.32 \\ \cline{4-8} 
                                            &                       &                       & $D^2$VINS (async) & 0.103          & 4.41           & 	0.088      & \textbf{0.21}          \\ \cline{4-8}
                                            &                       &                       & $D^2$VINS (single)& 0.139	         & 4.65           & 	0.161      & 0.31          \\ \cline{4-8}
                                            &                       &                       & VINS-Mono         & 0.235          & 6.08           & 0.264          & 0.52          \\ \hline\hline
    \multirow{2}{*}{Omni 5}                 & \multirow{3}{*}{45.4} & \multirow{3}{*}{1.03} & $D^2$VINS         & 0.075          & \textbf{0.71}  & \textbf{0.076} &\textbf{ 0.37} \\ \cline{4-8} 
                                            &                       &                       & $D^2$VINS (async) & \textbf{0.061} & 1.10           & 	0.078      & 0.38          \\ \cline{4-8}
                                            &                       &                       & $D^2$VINS (single)& 0.160          & 5.56           & 	0.225      & 0.47          \\ \cline{4-8}
                                            &                       &                       & VINS-Mono         & 0.282          & 4.21           & 	0.331      & 0.61          \\ \hline\hline
    \multirow{3}{*}{OmniLongNoYaw 5}        & \multirow{3}{*}{213.7}& \multirow{3}{*}{0.48} & $D^2$VINS         & 0.122          & 0.83           &  0.043         & 0.53 \\ \cline{4-8} 
                                            &                       &                       & $D^2$VINS (async) & \textbf{0.118} & \textbf{0.71}  & \textbf{0.031} & 0.51                \\ \cline{4-8} 
                                            &                       &                       & $D^2$VINS (single)& 0.150          & 4.36           & 0.109          & \textbf{0.39}          \\ \cline{4-8} 
                                            &                       &                       & VINS-Mono         & 0.500          & 1.78           & 	0.307      & 1.26          \\ \hline\hline
    \multirow{3}{*}{OmniLongYaw 5}          & \multirow{3}{*}{237.0}& \multirow{3}{*}{0.38} & $D^2$VINS         & 2.322          & 11.09          &  0.028         & \textbf{0.65} \\ \cline{4-8} 
                                            &                       &                       & $D^2$VINS (async) & 2.752          & 18.32          & \textbf{0.027} & 0.72          \\ \cline{4-8} 
                                            &                       &                       & $D^2$VINS (single)& \textbf{1.366} & \textbf{6.06}  & 1.019          & 1.22          \\ \cline{4-8} 
                                            &                       &                       & VINS-Mono         & 4.865          & 12.48          & 	1.466      & 3.40         \\ \hline\hline
                        \end{tabular}
\end{table*}

Table \ref{tab:tum} illustrates that our method achieves centimeter-level localization accuracy on the TUM ROOM dataset. Particularly during the first 30 seconds, where the common FoV is favorable, the relative localization accuracy is highly precise at 2.5cm. Even without common FoV for most of the evaluation, our dataset maintains similar centimeter-level accuracy as demonstrated in state-of-the-art works \cite{xu2020decentralized, xu2022omni}. This evaluation underscores $D^2$VINS's capability for accurate near-field state estimation using stereo cameras, especially with adequate common field of view, highlighting its potential for stereo camera-equipped UAV platforms.

We also test $D^2$VINS on custom datasets using omnidirectional cameras, with results detailed in Table \ref{tab:omni}. Our method consistently achieves centimeter-level relative localization accuracy across scenarios ranging from 2 to 5 UAVs. However, we observe a slight decrease in relative localization accuracy as the distance between UAVs increased, as indicated in Table \ref{tab:omni}. Despite this, the accuracy remains sufficient for inter-UAV collision avoidance. Notably, while both $D^2$VINS and VINS-Mono exhibit drift from ground truth over long flights with yaw rotation (as shown by the large ATE in Table \ref{tab:omni}), particularly in the OmniLongYaw 5 scenario, $D^2$VINS still maintains centimeter-level relative localization accuracy. This underscores its capability for reliable relative localization in extended flight durations.

In scenarios with sufficient common FoV, $D^2$VINS not only matches but often surpasses the state-of-the-art relative localization accuracy reported in \cite{xu2020decentralized, xu2022omni}, while also providing superior ego-motion estimation accuracy (evidenced by better ATE compared to VINS-Mono). In cases where common FoV is inconsistent, $D^2$VINS's relative localization accuracy is comparable to that in \cite{xu2020decentralized, xu2022omni}. Notably, our results outperform both VINS-Mono and $D^2$VINS (single) across almost all metrics in Tables \ref{tab:tum} and \ref{tab:omni}, demonstrating our algorithm's superiority in multi-robot scenarios. Specifically, the drift observed in VINS-Mono and $D^2$VINS (single) can significantly degrade their relative localization accuracy over extended periods, a drawback not evident in our approach.

Furthermore, Table \ref{tab:tum} and Table \ref{tab:omni} present accuracy comparisons of asynchronous $D^2$VINS (labeled as $D^2$VINS (async) in the tables). We observe that its performance closely parallels that of synchronous methods. However, in some instances, asynchronous $D^2$VINS exhibits slight instability in ATE. Given the communication benefits of asynchronous methods, opting for their deployment in real-world environments represents a reasonable compromise.

\subsection{Evaluation of $D^2$PGO}

\begin{table*}[h]
    \centering
    \caption{\small{This table compares $D^2$PGO with and without rotation initialization, detailing both initial and final costs, as well as the size of the pose graph, which includes the counts of keyframes and edges.}}
    \label{tab:d2pgo}
    \begin{tabular}{c|c|c|c|c|c|c|c|c}
    \hline\hline
    Dataset                             & Avg. Traj. Len.               &  Keyframes              & Edges                  & Method                &  Initial Cost                & Final Cost    & Iterations & Solve Time (s) \\ \hline\hline
    \multirow{2}{*}{TUM Corr 5}         & \multirow{2}{*}{255.6}        & \multirow{2}{*}{12065} & \multirow{2}{*}{16814} & $D^2$PGO              &  \multirow{2}{*}{350408.6}   &  61.2         &   234.2    &  10.1          \\ \cline{5-5}\cline{7-9} 
                                        &                               &                                     &                                     & $D^2$PGO (NoRotInit)  & 	                           & 1920.4        &   238.4    &  10.1          \\ \hline\hline
    \multirow{2}{*}{OmniLongYaw 5}      & \multirow{2}{*}{242.6}        & \multirow{2}{*}{16430} & \multirow{2}{*}{51358} & $D^2$PGO              &  \multirow{2}{*}{520424.1}   &  1366.3       &   37.2     &  20.5          \\ \cline{5-5}\cline{7-9} 
                                        &                               &                                     &                                     & $D^2$PGO (NoRotInit)  & 	                           & 41072.3       &   31.6     &  20.4          \\ \hline\hline  
    \multirow{2}{*}{HKUST RI 3}         & \multirow{2}{*}{164.5}        & \multirow{2}{*}{3347}  & \multirow{2}{*}{3677}  & $D^2$PGO              &  \multirow{2}{*}{6438.5}     &  12.1         &   152.0    &  3.0           \\ \cline{5-5}\cline{7-9} 
                                        &                               &                                     &                                     & $D^2$PGO (NoRotInit)  & 	                           &  1301.8       &   197.0    &  3.0           \\ \hline\hline  
    \end{tabular}
\end{table*}

Table \ref{tab:d2pgo} showcases $D^2$PGO's optimization results for various pose graphs generated by the front-ends of $D^2$SLAM and $D^2$VINS. During the evaluation, we incorporated a consistent 50ms communication delay to emulate real-world conditions. The comparison of $D^2$PGO's performance, with and without rotation initialization, reveals that rotation initialization plays a crucial role in preventing $D^2$PGO from converging to local optima.

\subsection{Evaluation of $D^2$SLAM}

\begin{table}[h]
    \centering
    \caption{\small{The table presents statistical results comparing $D^2$SLAM with DOOR-SLAM \cite{lajoie2020door} on our custom datasets. The estimated trajectories and ground truth are aligned using multiple states \cite{Zhang18iros}, based on the first robot's estimations.
    }}
    \label{tab:d2slam}
    \resizebox{1.0\linewidth}{!}{
    \begin{tabular}{c|c|c|c|c|c}
    \hline\hline
    Dataset                                 & Method            & $ATE_{pos}$    & $ATE_{rot}$   & $RE_{pos}$      & $RE_{rot}$     \\ \hline\hline
    \multirow{2}{*}{OmniLongYaw 5}          & $D^2$SLAM         & \textbf{0.168} & 3.33          &  \textbf{0.037} & 0.91 \\ \cline{2-6} 
                                            & DOOR-SLAM         & 0.149          & \textbf{3.47} & 	0.085          & \textbf{0.72}          \\ \hline\hline 
    \multirow{2}{*}{TUM ROOM 5}             & $D^2$SLAM         & \textbf{0.032} & \textbf{0.72} &  \textbf{0.079} & 1.90          \\ \cline{2-6} 
                                            & DOOR-SLAM         & 0.064          & 1.47          & 	0.122          & \textbf{1.86} \\ \hline\hline 
    \multirow{2}{*}{TUM Corr 5}             & $D^2$SLAM         & 0.275          & 3.93          &  0.206          & \textbf{1.77} \\ \cline{2-6} 
                                            & DOOR-SLAM         & \textbf{0.068} & \textbf{1.26} & 	\textbf{0.145} & 2.39           \\ \hline\hline 
    
    \end{tabular}}
    \vspace{-0.5cm}
\end{table}
\begin{figure}[ht]
    \centering
    \includegraphics[width=\linewidth]{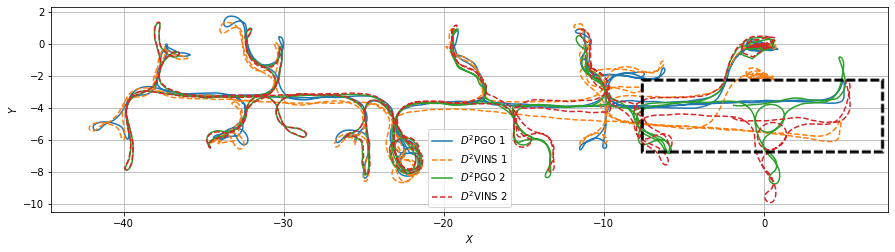}
    \caption{\small{The figure illustrates $D^2$SLAM's estimated trajectories on the TUM Corr 2 Dataset, a combination of TUM VI corridor datasets {1, 2}. In this dataset, two robots sequentially traverse the same corridor, marked by the black box section. Within this section, the trajectories from $D^2$VINS exhibit significant drift, whereas those from $D^2$PGO show minimal drift and maintain good alignment between the two robots.
    }}\label{fig:$D^2$PGO}
    \vspace{-0.5cm}
\end{figure}

\begin{figure*}[h]
    \centering
    \includegraphics[width=1.0\linewidth]{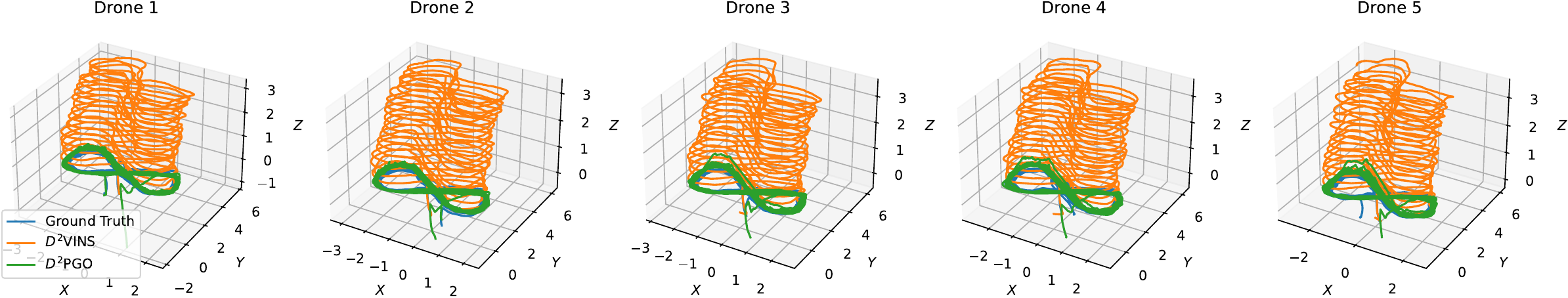}
    \caption{
        \small{The images display trajectories estimated by $D^2$VINS and $D^2$PGO in the OmniLongYaw 5 dataset. The trajectories from $D^2$VINS show noticeable drift, whereas those from $D^2$PGO remain closely aligned with the ground truth.}
    }\label{fig:OmniLongYaw5}
    \vspace{-0.5cm}
\end{figure*}

We further evaluated $D^2$SLAM on both public and custom datasets to assess the performance of our near- and far-field state estimation. Leveraging the high accuracy of VIO's angle measurement and aiming for optimal real-time performance, only the second stage of $D^2$PGO was utilized in these tests. We conducted a comparison between $D^2$SLAM and the state-of-the-art distributed CSLAM system DOOR-SLAM \cite{lajoie2020door} using datasets with ground truth, with results presented in Table \ref{tab:d2slam}. For this comparison, we applied our custom implementation in $D^2$SLAM to generate PGO for DOOR-SLAM, ensuring consistency and eliminating potential discrepancies from different frontends.

Table \ref{tab:d2slam} shows that our method outperforms DOOR-SLAM in relative state estimation on the first two datasets. This advantage arises because DOOR-SLAM relies on pose graph optimization, a common but loosely coupled approach in CSLAM. While both methods demonstrate comparable global consistency accuracy due to similar inputs, DOOR-SLAM exhibits superior performance on the TUM Corr dataset. 
This can be attributed to the partial ground truth available only at the beginning and end of the TUM Corr dataset, coupled with inadequate yaw overlap between UAVs towards the dataset's end.

We conduct further tests of $D^2$SLAM on larger-scale datasets, including TUM VI Corr 5, HKUST RI 3, and OmniLongYaw 5, to evaluate its global consistency. The real-time odometry and final pose graphs estimated in these experiments are illustrated in Fig. \ref{fig:$D^2$PGO}, Fig. \ref{fig:ri_2_traj}, and Fig. \ref{fig:OmniLongYaw5}. The figures reveal a stark contrast: while trajectories from $D^2$VINS diverge from the starting point, those estimated by $D^2$PGO consistently return to the start. Notably, in Fig. \ref{fig:$D^2$PGO}, $D^2$PGO accurately aligns the trajectories of two robots in the corridor (highlighted by the black box), unlike $D^2$VINS, where drift occurs due to the robots not traversing the corridor simultaneously. These results affirm the global consistency of $D^2$SLAM in far-field scenarios.
\begin{figure*}[ht!]
    \centering
    \settowidth\aimage{\includegraphics[height=3cm]{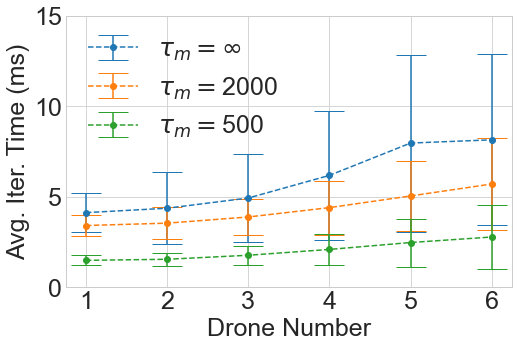}}
    \settowidth\bimage{\includegraphics[height=3cm]{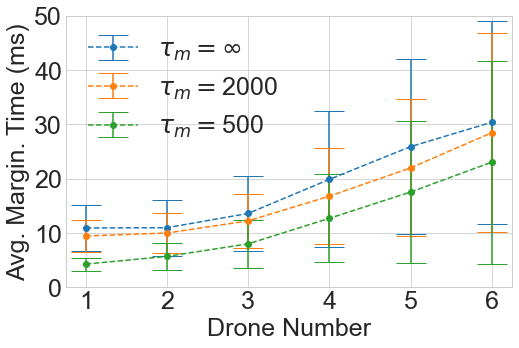}}
    \settowidth\cimage{\includegraphics[height=3cm]{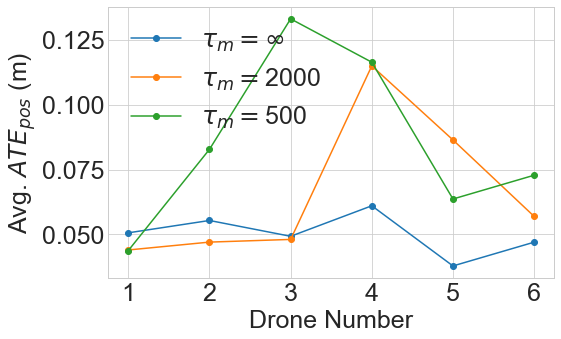}}
    \settowidth\dimage{\includegraphics[height=3cm]{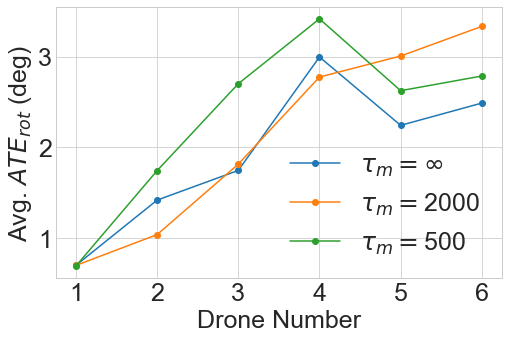}}
    \resizebox{\linewidth}{!}{
    \begin{tabular}{p{\aimage-\tabcolsep}p{\bimage-\tabcolsep}p{\cimage-\tabcolsep}p{\dimage-\tabcolsep}}
        \includegraphics[height=3cm]{scability_stereo_iter_time}\newline\vspace{-0.5cm}
        \subcaption{}\label{fig:scability_stereo_iter_time} &   
        \includegraphics[height=3cm]{scability_stereo_marginalization_time}\newline\vspace{-0.5cm}
        \subcaption{}\label{fig:scability_stereo_marginalization_time} &  
        \includegraphics[height=3cm]{scability_stereo_ATE_trans}\newline\vspace{-0.5cm}
        \subcaption{}\label{fig:scability_stereo_ATE_trans} &    
        \includegraphics[height=3cm]{scability_stereo_ATE_rot}\newline\vspace{-0.5cm}
        \subcaption{}\label{fig:scability_stereo_ATE_rot}    
    \end{tabular}
    }

    \settowidth\aimage{\includegraphics[height=3cm]{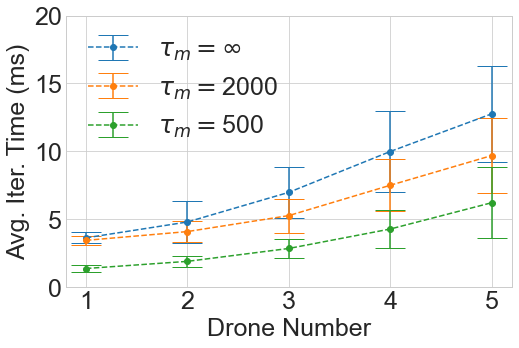}}
    \settowidth\bimage{\includegraphics[height=3cm]{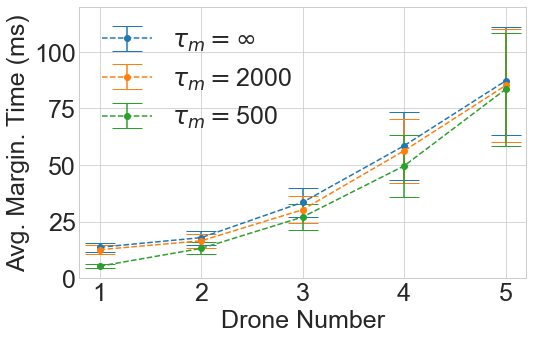}}
    \settowidth\cimage{\includegraphics[height=3cm]{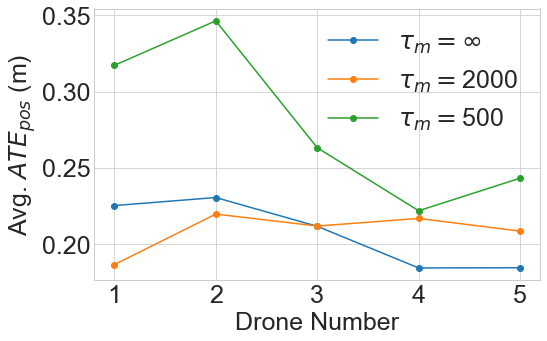}}
    \settowidth\dimage{\includegraphics[height=3cm]{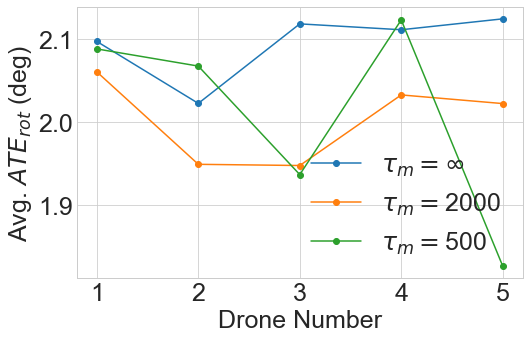}}
    \resizebox{\linewidth}{!}{
    \begin{tabular}{p{\aimage-\tabcolsep}p{\bimage-\tabcolsep}p{\cimage-\tabcolsep}p{\dimage-\tabcolsep}}
        \includegraphics[height=3cm]{scability_omni_iter_time}\newline\vspace{-0.5cm}
        \subcaption{}\label{fig:scability_omni_iter_time} &   
        \includegraphics[height=3cm]{scability_omni_marginalization_time}\newline\vspace{-0.5cm}
        \subcaption{}\label{fig:scability_omni_marginalization_time} &  
        \includegraphics[height=3cm]{scability_omni_ATE_trans}\newline\vspace{-0.5cm}
        \subcaption{}\label{fig:scability_omni_ATE_trans} &    
        \includegraphics[height=3cm]{scability_omni_ATE_rot}\newline\vspace{-0.5cm}
        \subcaption{}\label{fig:scability_omni_ATE_rot}    
    \end{tabular}
    }
    \vspace{-0.5cm}
    \caption{\small{
        The image illustrates the average time per iteration, marginalization time, translation error, and rotation error of $D^2$VINS for varying numbers of robots. We set $\tau_l=200$ for the evaluation and compared results across different $\tau_m$ values.
        Sub-figures a)-d) represent stereo camera results derived from the TUM Room 5 dataset, while e)-h) show results from quad fisheye cameras using the OmniLongYaw 5 dataset.
       }}\label{fig:scability}
\end{figure*}

\begin{figure}[h!]
    \centering
    \settowidth\limage{\includegraphics[height=3cm]{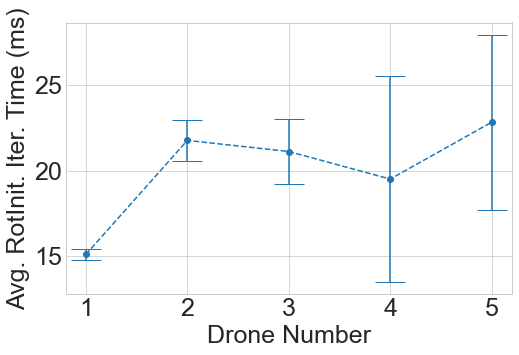}}
    \settowidth\rimage{\includegraphics[height=3cm]{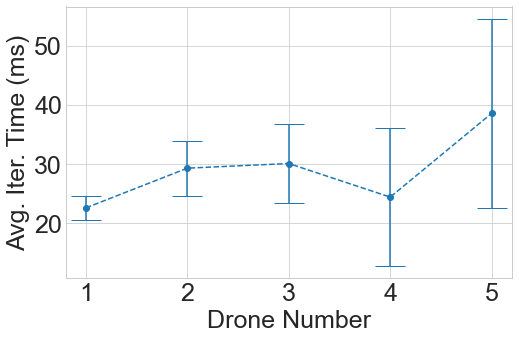}}
    \resizebox{1.0\linewidth}{!}{
    \begin{tabular}{p{\limage-\tabcolsep}p{\rimage-\tabcolsep}}
        \includegraphics[height=3cm]{scability_pgo_rotinit_time}\newline\vspace{-0.5cm} \subcaption{}\label{fig:scability_pgo_rotinit_time} &  
        \includegraphics[height=3cm]{scability_pgo_arock_iter_time}\newline\vspace{-0.5cm} \subcaption{}\label{fig:scability_pgo_arock_iter_time}
    \end{tabular}
    }
    \caption{\small{
        The image depicts the average iteration time of $D^2$PGO with varying numbers of robots, derived from the TUM Corr 5 dataset. Sub-figure a) presents the results of rotation initialization, while e)-h) show the results of ARockPGO.       }}\label{fig:scability_pgo}
        \vspace{-0.5cm}
\end{figure}

\begin{table*}[h!]
    \centering
    \caption{Typical computational time and running frequency of each module in the real-world experiment of $D^2$VINS.}
    \label{tab:comp_time}
    \begin{tabular}{c|c|c|c|c|c}
    \hline\hline
    Item      & \begin{tabular}[c]{@{}c@{}}Image \\ Preprocesing \end{tabular} & \begin{tabular}[c]{@{}c@{}}Feature Tracking \\ (Local)\end{tabular} & \begin{tabular}[c]{@{}c@{}}Feature Tracking \\ (Remote)\end{tabular} & Optimization & Marginalization \\ \hline\hline
    Frequency      & 15 Hz                                                         & 15 Hz                                                               & 5 Hz                                                                 & 5 Hz         & 5 Hz            \\ \hline
    Time cost & 41.7ms                                                        & 2.67ms                                                              & 1.64ms                                                               & 56.23ms      & 16.88ms         \\ \hline\hline
    \end{tabular}
    \vspace{-0.5cm}
\end{table*}
Fig. \ref{fig:dense_mapping_2} displays the results of TSDF reconstruction using $D^2$PGO combined with submap fusion. The absence of noticeable drift in the map further also confirms the global consistency of $D^2$SLAM.

\subsection{Scalability}

In this section, we explore the scalability of computational performance.

\subsubsection{Front-end}

The computational demands of remote feature matching and loop closure detection in the front-end of $D^2$SLAM scale approximately linearly with the number of nearby robots. These algorithms are efficient, and their performance in real-world experiments is detailed in Table \ref{tab:comp_time}.

\subsubsection{Back-end}
The back-end dynamics are more complex. We adopt distributed optimization methods for graph optimization in this paper. In practice, $D^2$VINS implements a tailored landmark selection algorithm, MLS, balancing computational efficiency, accuracy, and robustness. This algorithm allows the number of visual measurements processed by $D^2$VINS to scale with the increase in robots, up to the $\tau_m$ limit. We have chosen a relatively high $\tau_m$ value to enhance the robustness of $D^2$VINS.

In Figs. \ref{fig:scability} and \ref{fig:scability_pgo}, we demonstrate the scalability of $D^2$VINS and DPGO by charting the average computation time per operation against the increasing swarm size. Notably, on the TUM Room 5 dataset, there is a slight decrease in rotation accuracy with more robots, likely due to reduced inter-robot overlap in stereo camera FoVs. This issue doesn't occur in the OmniLongYaw 5 dataset with omnidirectional cameras. As the swarm size expands, the problem size grows quadratically. However, $D^2$SLAM's computational requirement for optimization increases at no more than a linear rate, showcasing the algorithm's efficiency. Furthermore, $D^2$SLAM provides flexibility in managing computational load through adjustable parameters $\tau_m$ and $\tau_l$, suitable for diverse computational platforms. Reducing computational complexity through these parameters doesn't significantly compromise accuracy.

The time taken for marginalization increases with the number of robots, although, in theory, marginalization computation should be similar to a single optimization linear iteration. This discrepancy is primarily due to the ceres-solver framework's conventional approach of converting marginalization results into square root form for optimization \cite{qin2017vins,leutenegger2015keyframe}, which we plan to further optimize in the future.

\begin{table*}[]
    \centering
    \caption{Typical message sizes and broadcast frequencies in $D^2$SLAM. The terms 'D2VINS Update' and 'D2PGO Update' in the table refer to messages used for synchronization in swarm optimization within $D^2$SLAM. Here, $p^k$ represents the number of poses in each UAV's sliding window, and $e^k$ denotes the sum of dual states in the ARock algorithm.}
    \label{tab:comm}
    \begin{tabular}{c|c|c|c|c|c|c}
    \hline\hline
    Item      & \begin{tabular}[c]{@{}c@{}}Complete Keyframe\\ (Stereo)\end{tabular} & \begin{tabular}[c]{@{}c@{}}Complete Keyframe\\ (Omni)\end{tabular} & \begin{tabular}[c]{@{}c@{}}Compact Keyframe\\ (Stereo)\end{tabular} & \begin{tabular}[c]{@{}c@{}}Compact Keyframe\\ (Omni)\end{tabular} & $D^2$VINS Update & $D^2$PGO Update                           \\ \hline\hline
    Size      & 20.1kB                                                               & 56.3kB                                                             & 1.2kB                                                               & 4.8kB                                                             & $0.188+0.036p^k$kB &$0.056+0.064e^k$kB \\ \hline
    Frequency & 5                                                                    & 5                                                                  & 5                                                                   & 5                                                                 & 5                & 1                                         \\ \hline\hline
    \end{tabular}
    \vspace{-0.5cm}
    \end{table*}

    \begin{table}[h]
        \centering
        \caption{\small{The total communication volume in MB of $D^2$SLAM in baseline mode and $D^2$SLAM mode.}}
        \label{tab:d2slam_comm}
        \begin{tabular}{c|c|c|c|c}
        \hline\hline
        \multirow{2}{*}{Dataset}                & \multirow{2}{*}{Mode} & \multirow{2}{*}{Front-end}    & \multicolumn{2}{c}{Back-end}                      \\\cline{4-5}
                                                &                       &                               & $D^2$VINS                         & $D^2$PGO      \\ \hline\hline      
        \multirow{2}{*}{OmniLongYaw 5}          & Baseline              &     1546.7                    &   75.9                            &  2340.9       \\ \cline{2-5} 
                                                & $D^2$SLAM             &      984.2                    &  70.8                             &  2179.4       \\ \hline\hline 
        \multirow{2}{*}{TUM ROOM 5}             &  Baseline             &     266.4                    &  1.6                               &  42.0         \\ \cline{2-5} 
                                                & $D^2$SLAM             &    244.2                      &  1.2                              & 31.0       \\ \hline\hline 
        \multirow{2}{*}{TUM Corr 5}             &  Baseline             & 344.2                         &  2.4                              & 60.1         \\ \cline{2-5} 
                                                & $D^2$SLAM             & 323.0                         &  2.9                              & 74.1          \\ \hline\hline 
        \multirow{2}{*}{HKUST RI 3}             &  Baseline             & 231.8                         &  10.1                             & 13.4          \\ \cline{2-5} 
                                                & $D^2$SLAM             & 105.4                         &  9.6                              & 11.9          \\ \hline\hline 
        \end{tabular}
    \end{table}
    \vspace{-0.5cm}
\subsection{Communication}

Effective communication is pivotal for successful aerial swarm operations. To shed light on this aspect, Table \ref{tab:comm} details the sizes and broadcast frequencies of key messages transmitted by $D^2$SLAM on each UAV. 
Additionally, Table \ref{tab:d2slam_comm} presents a comparison between $D^2$SLAM's communication strategy (as detailed in Sect. \ref{sect:comm_strategy}) and a  baseline method. This baseline strategy involves broadcasting all measurements (represented as complete keyframes in our context) without changing communication modes and is referenced in \cite{xu2022omni}. $D^2$SLAM's tailored strategy significantly reduces communication volume, especially when UAVs are widely dispersed.

Transitioning from the broader communication strategy, we delve into specifics for a 5-UAV swarm. Here, the maximum value of $p^k$ reaches 51, leading to a maximum $D^2$VINS update size of 3.5kB per timestep. The value of $e^k$, which varies depending on the map and environment. For example, in TUM Corr 5 dataset, a typical $e^k$ value is 1262, resulting in each $D^2$PGO update being approximately 81kB.

\subsection{Real-world Experiments of $D^2$SLAM}

\begin{figure}[ht]
    \centering
    \includegraphics[width=0.9\linewidth]{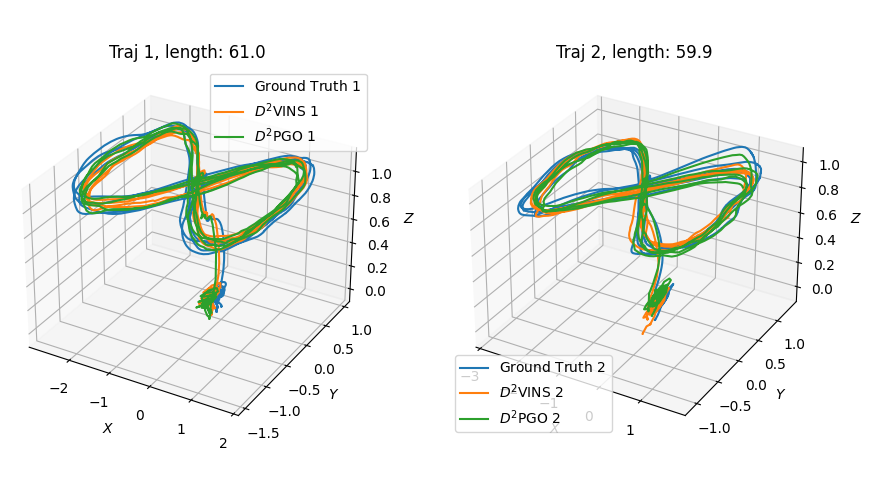}
    \caption{The images display trajectories estimated by $D^2$VINS and $D^2$PGO in a real-world experiment.}\label{fig:realworld-2}
    \vspace{-0.6cm}
\end{figure}

Beyond dataset validation, we also test $D^2$SLAM in realistic environments using our UAV platforms. In these real-world experiments, as depicted in Fig. \ref{fig:realworld-2} and Table \ref{tab:realworld-2}, $D^2$SLAM successfully handles limitations in computing power and communication, performing as anticipated. The table includes data from two flights, conducted by UAVs equipped with Realsense d435i cameras and utilizing the stereo version of $D^2$SLAM. In the \textit{UAV2 NoYaw} experiment, the UAVs maintain fixed yaw angles, whereas in \textit{UAV2 Yaw}, they rotate yaw angles according to their flight paths.

\begin{table*}[h!]
    \centering
    \caption{The table presents $D^2$SLAM's accuracy results from two flight experiments. The alignment approach for evaluating accuracy is consistent with that used in Table \ref{tab:d2slam}.}
    \label{tab:realworld-2}
    \begin{tabular}{c|c|cc|cc|ccc|c}
    \hline\hline
                                 &                                   & \multicolumn{2}{c|}{$ATE_{pos}$}          & \multicolumn{2}{c|}{$ATE_{rot}$}          & \multicolumn{3}{c|}{$RE_{pos}$}                                 &                              \\ \cline{3-9}
    \multirow{-2}{*}{Experiment} & \multirow{-2}{*}{Avg. Traj. Len.} & \multicolumn{1}{c|}{$D^2$VINS} & $D^2$PGO & \multicolumn{1}{c|}{$D^2$VINS} & $D^2$PGO & \multicolumn{1}{c|}{x}     & \multicolumn{1}{l|}{y}     & z     & \multirow{-2}{*}{$RE_{rot}$} \\ \hline\hline
    UAV2 NoYaw                 & 51.0                              & \multicolumn{1}{c|}{0.103}     & 0.090    & \multicolumn{1}{c|}{2.24°}     & 1.95°    & \multicolumn{1}{c|}{0.047} & \multicolumn{1}{l|}{0.067} & 0.056 & {\color[HTML]{000000} 3.36°} \\ \hline
    UAV2 Yaw                   & 73.3                              & \multicolumn{1}{c|}{0.219}     & 0.147    & \multicolumn{1}{c|}{5.69°}     & 4.17°    & \multicolumn{1}{c|}{0.124} & \multicolumn{1}{l|}{0.144} & 0.100 & 8.88°                        \\ \hline\hline
    \end{tabular}
    \vspace{-0.5cm}
\end{table*}

$D^2$SLAM demonstrates robust global consistency in both scenarios, with the ATE of $D^2$PGO being only about ten centimeters for flights ranging between 50 and 70 meters. This ATE accuracy of $D^2$PGO surpasses that of $D^2$VINS, highlighting the crucial role of $D^2$PGO in achieving global consistency for $D^2$SLAM. In terms of relative localization, $D^2$SLAM achieves centimeter-level accuracy with fixed yaw. However, the rotating yaw experiment reveals reduced relative localization accuracy, a consequence of the stereo version's limited FoV, as previously discussed.

Table \ref{tab:comp_time} displays the performance and operating frequency of key $D^2$SLAM algorithms when executed on the onboard computer during real-world experiments. Furthermore, communication between UAVs in these experiments is smooth: the average communication delay is only 23.89ms. This demonstrates that $D^2$SLAM meets real-time operational requirements in real-world aerial swarm systems.

\section{Conclusion and Future Work}\label{sect:con}
In this paper, we introduce $D^2$SLAM, a distributed and decentralized collaborative visual-inertial SLAM system. Our experiments demonstrate that $D^2$SLAM excels in real-time, high-precision local localization, and maintains high-precision relative localization when UAVs are in proximity, showcasing its capability for accurate near-field state estimation. Additionally, $D^2$SLAM can simultaneously estimate globally consistent trajectories, ensuring far-field state estimation with global consistency. With its flexible sensor configuration and versatile state estimation capabilities, $D^2$SLAM is poised to advance aerial swarm research, with applications ranging from self-assembling aerial swarms and cooperative transportation to inter-UAV collision avoidance and exploration of unknown environments.

While $D^2$SLAM, as presented in this paper, demonstrates considerable potential, it also has areas for future improvement:
1) Despite its distributed backend, the scalability of swarm size is currently constrained by communication and front-end computing capabilities. Future enhancements to $D^2$SLAM aim to enable its application to significantly larger aerial swarms.
2) Unlike our previous work, $D^2$SLAM adopts a more traditional visual SLAM approach without incorporating relative measurements such as UWB or mutual visual detection. This design choice makes $D^2$SLAM adaptable to a wider range of scenarios, including those with occlusions affecting UWB measurements or challenges in accurate visual identification of robots. However, recognizing the efficiency of relative measurements in certain contexts, we plan to integrate these into $D^2$SLAM in future developments.
3) Future improvement will also investigate leveraging hardware layer information or dense map data to assess communication quality between UAVs. This initiative aims to optimize network bandwidth utilization for state estimation algorithms.

Furthermore, we envision $D^2$SLAM's applicability extending beyond multi-robot systems to serve as a distributed visual-inertial SLAM system within a single robot. For example, deploying multiple $D^2$SLAM nodes on a single rigid UAV or ground robot could enhance state estimation accuracy and resilience against individual node failures. These nodes could also form distributed stereo camera systems to extend the range of environmental perception, as explored in \cite{karrer2021distributed}. Additionally, deploying $D^2$SLAM on multi-rigid robots, such as multi-legged platforms, and incorporating specific constraints could further refine their state and environmental estimations while offering redundancy.

\newlength{\bibitemsep}\setlength{\bibitemsep}{.03\baselineskip}
\newlength{\bibparskip}\setlength{\bibparskip}{0pt}
\let\oldthebibliography\thebibliography
\renewcommand\thebibliography[1]{%
 \oldthebibliography{#1}%
 \setlength{\parskip}{\bibitemsep}%
 \setlength{\itemsep}{\bibparskip}%
}
\bibliographystyle{IEEEtran}
\bibliography{hao2022.bib}

\begin{IEEEbiography}[{\includegraphics[width=1in,height=1in,clip,keepaspectratio]{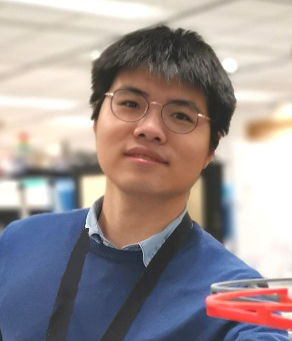}}]{Hao Xu}
    received the B.Sc. degree in Physics from the University of Science and Technology, and the Ph.D. degree in Electronic and Computer Engineering (ECE) from the Hong Kong University of Science and Technology, Hong Kong, in 2023. He is currently working as an algorithm engineer at Shenzhen DJI Sciences and Technologies Ltd. His research interests include unmanned aerial vehicles, aerial swarm, state estimation, sensor fusion, localization, and mapping.\end{IEEEbiography}
\vspace{-1cm}
\begin{IEEEbiography}[{\includegraphics[width=1in,height=1in,clip,keepaspectratio]{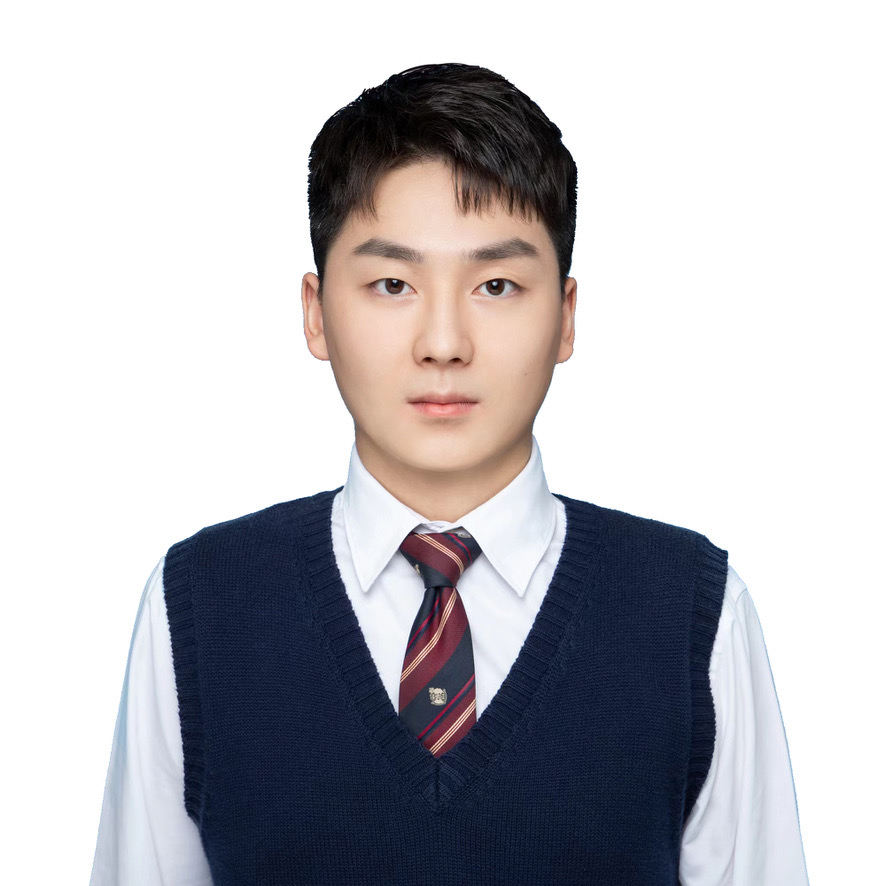}}]{Peize LIU} received the B.Sc. degree in Software Engineering  from
    the University of Electronic Science and Technology of China,
    Chengdu, China, in 2022. He is currently working
    toward the Ph.D. degree with the Hong Kong University of Science and Technology, Hong Kong, under
    the supervision of Prof. Shaojie Shen. His research interests include unmanned aerial vehicles, state estimation in aerial swarm, and swarm system.\end{IEEEbiography}
\vspace{-1cm}
\begin{IEEEbiography}[{\includegraphics[width=0.75in,height=1in,clip,keepaspectratio]{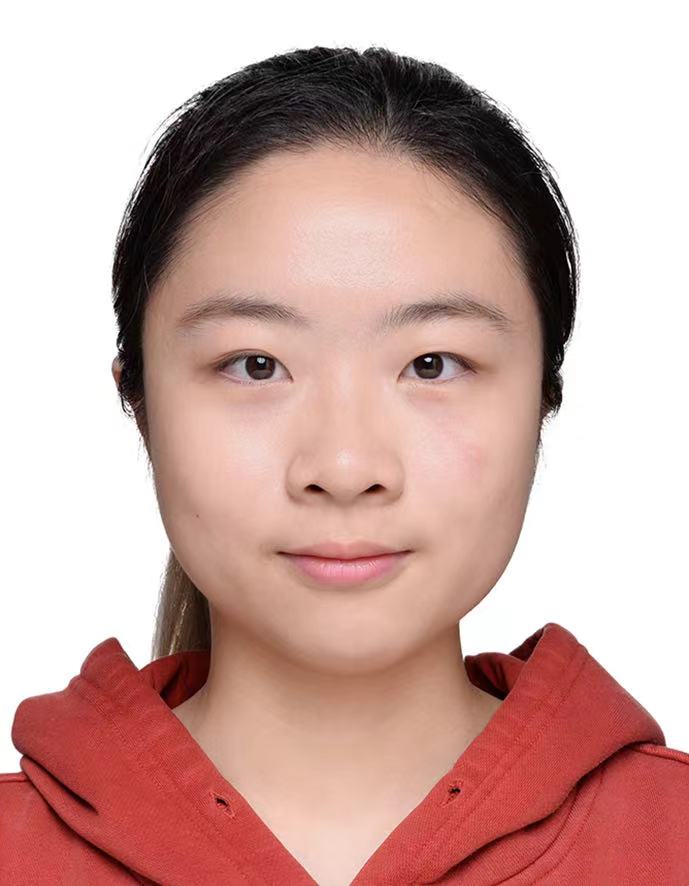}}]{Xinyi Chen}
     received the B.Sc. degree in Mathematics, and in Computer Science from the Hong Kong University of Science and Technology, Hong Kong, in 2021. She is currently working towards the Ph.D. degree with the Hong Kong University of Science and Technology, Hong Kong, under the supervision of Prof. Shaojie Shen.
    Her research interests include aerial vehicles, motion planning, autonomous exploration, aerial swarm, localization and mapping.\end{IEEEbiography}
\vspace{-1cm}
\begin{IEEEbiography}[{\includegraphics[width=0.75in,height=1in,clip,keepaspectratio]{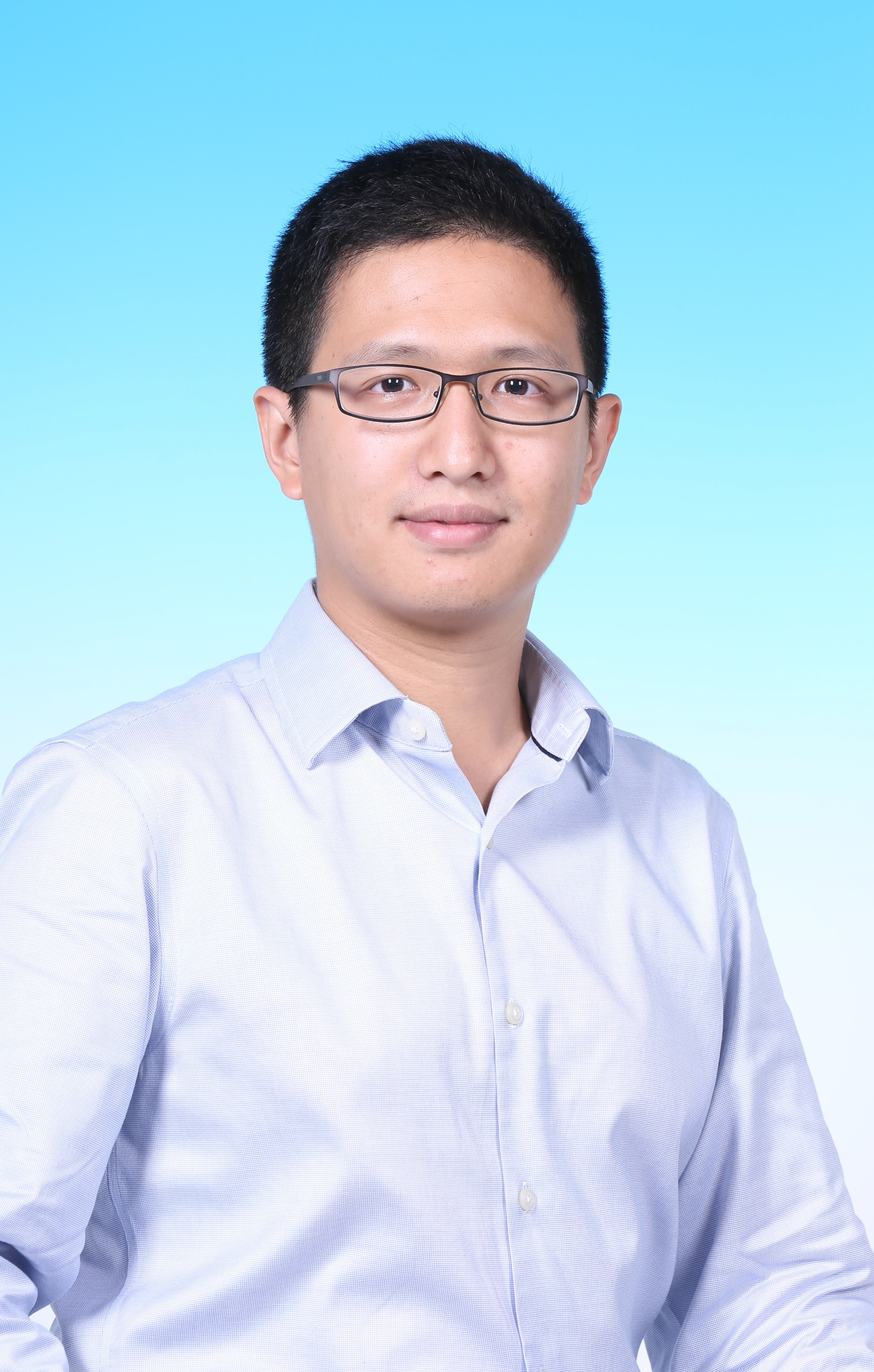}}]{Shaojie Shen} received his B.Eng. degree in Electronic Engineering from the Hong Kong University of Science and Technology (HKUST) in 2009. He received his M.S. in Robotics and Ph.D. in Electrical and Systems Engineering in 2011 and 2014, respectively, all from the University of Pennsylvania.
    He joined the Department of Electronic and Computer Engineering at the HKUST in September 2014 as an Assistant Professor, and is promoted to Associate Professor in July 2020. He is the founding director of the HKUST-DJI Joint Innovation Laboratory (HDJI Lab). His research interests are in the areas of robotics and unmanned aerial vehicles, with focus on state estimation, sensor fusion, localization and mapping, and autonomous navigation in complex environments. \end{IEEEbiography}

\end{document}